\pdfoutput=1
\documentclass[11pt]{article}

\usepackage{ACL2023}
\usepackage{times}
\usepackage{latexsym}
\usepackage[T1]{fontenc}
\usepackage[utf8]{inputenc}
\usepackage{microtype}
\usepackage{inconsolata}
\usepackage{xcolor}
\usepackage{adjustbox}
\usepackage{diagbox}
\usepackage{amsmath,amssymb}
\usepackage[inline]{enumitem}
\usepackage{array,multirow,graphicx,makecell}
\usepackage{tabularx}
\usepackage{float}
\usepackage{booktabs}
\newcommand{\com}[1]{}

\newcolumntype{C}[1]{>{\centering\arraybackslash}m{#1}}

\newcommand{\xsum}{\texttt{XSUM40}}
\newcommand{\squad}{\texttt{SQuAD17}}
\newcommand{\art}{\texttt{ART10}}
\newcommand{\shakespeare}{\texttt{Shake7}}

\newcommand{\bartl}{\texttt{BART-L}}
\newcommand{\bartb}{\texttt{BART-6:6}}

\newcommand{\barts}{\texttt{BART-6:2}}
\newcommand{\bartp}{\texttt{BART-2:6}}
\newcommand{\bartkd}{\texttt{BART-KD}}
\newcommand{\tfivel}{\texttt{T5-L}}
\newcommand{\tfives}{\texttt{T5-S}}
\newcommand{\tfivekd}{\texttt{T5-KD}}
\newcommand{\gpts}{\texttt{GPT2}}
\newcommand{\gptm}{\texttt{GPT2-M}}
\newcommand{\gptl}{\texttt{GPT2-L}}
\newcommand{\opts}{\texttt{OPT-125M}}
\newcommand{\optm}{\texttt{OPT-350M}}

\newcommand{\bleu}{\textsc{BLEU}}
\newcommand{\rouge}{\textsc{ROUGE}}

\newcommand{\bertscore}{\textsc{BS}}
\newcommand{\ppl}{\textsc{PPL}}

\newcommand{\TBD}[1]{\textcolor{red}{TBD}}

\definecolor{mred}{RGB}{192, 0, 0}
\definecolor{mgreen}{RGB}{112, 173, 71}

\title{A Systematic Study of Knowledge Distillation for Natural Language Generation with Pseudo-Target Training}

\author{Nitay Calderon\thanks{\hspace{0.25cm} Work was mainly done during an internship at Microsoft MSAI. Contact: \texttt{\href{mailto:nitay@campus.technion.ac.il}{nitay@campus.technion.ac.il}}. Code: \texttt{\href{https://github.com/nitaytech/KD4Gen}{https://github.com/nitaytech/KD4Gen}}.} \\ 
 Technion -- IIT
        \And  
        Subhabrata Mukherjee \\ Microsoft Research
        \And  
        Roi Reichart \\ Technion -- IIT
        \And  
        Amir Kantor \\ Microsoft}

\date{}

\begin{document}
\maketitle
\begin{abstract}

Modern Natural Language Generation (NLG) models come with massive computational and storage requirements. In this work, we study the potential of compressing them, which is crucial for real-world applications serving millions of users. We focus on Knowledge Distillation (KD) techniques, in which a small student model learns to imitate a large teacher model, allowing to transfer knowledge from the teacher to the student. In contrast to much of the previous work, our goal is to optimize the model for a specific NLG task and a specific dataset. Typically in real-world applications, in addition to labeled data there is abundant unlabeled task-specific data, which is crucial for attaining high compression rates via KD. In this work, we conduct a systematic study of task-specific KD techniques for various NLG tasks under realistic assumptions. We discuss the special characteristics of NLG distillation and particularly the exposure bias problem. Following, we derive a family of Pseudo-Target (PT) augmentation methods, substantially extending prior work on sequence-level KD. We propose the Joint-Teaching method, which applies word-level KD to multiple PTs generated by both the teacher and the student.
Finally, we validate our findings in an extreme setup with no labeled examples using GPT-4 as the teacher. 
Our study provides practical model design observations and demonstrates the effectiveness of PT training for task-specific KD in NLG.

\end{abstract}
\section{Introduction}
\label{sec:intro}

Modern \emph{Natural Language Generation (NLG)} systems are based on \emph{pre-trained Language Models (LMs)}, which are gradually achieving remarkable milestones \citep{t5, gpt3, gpt-4}.
Alongside the impressive advances in applications such as Neural Machine Translation (NMT), Summarization, chatbots, such models have also become increasingly larger, deeper, slower, and more complex. The massive storage requirements and high computational complexity of NLG models discourage their deployment in real-life. As such, there is a growing demand in the industry for compressing such models while preserving their performance. 

Model compression methods typically either prune less informative parameters \citep{first_pruning} or use \emph{knowledge distillation (KD)} \citep{first_kd, sequence_level} to transfer knowledge from a larger model (the teacher) to a smaller model (the student). In generation tasks, KD can be applied at the word-level, by training the student to mimic the teacher's next token distribution, or at the sequence-level, by training the student on \emph{Pseudo-Targets (PTs)} generated by the teacher. 

Although KD research is extensive \citep{kd_cv_survey, compression_survey, efficient_survey, kd_survey_xu}, most works focus on Natural Language Understanding (NLU) tasks, task-agnostic language modeling, or specific generation tasks (e.g., NMT). 
Additionally, KD works for NLG typically consider large datasets with hundreds of thousands of labeled examples, and ignore unlabeled data \citep{kd_summarization,   selective_kd, quant_sum, attention_pts}.

In more realistic scenarios, however, the number of labeled examples is limited, alongside an abundance of unlabeled data \citep{realistic_ssl, docogen} that may contribute to KD. To bridge these gaps, in this paper we conduct a systematic study of KD for NLG, considering a variety of tasks: Summarization, Question Generation, Abductive Reasoning, Style Transfer and Simplification, in a more realistic setup.

Our realistic setup follows 5 criteria that are particularly attractive for a broad range of NLP practitioners: (1) Only several thousand labeled examples are available for training (Medium-resource), as annotation is costly or labor-intensive, especially for NLG. This is in contrast to research setups where labeled datasets can be very large. 
(2) Large amounts of unlabeled data are available, as is often the case in industrial setups where unlabeled data is collected during the life-cycle of the product; (3) Off-the-shelf models are used, which is more practical than training models from scratch; (4) Inference-time efficiency is our goal, meaning high compression rate; (5) One-time computational training resources are negligible, compared to inference-time, allowing extensive use of PTs.

Recently, huge LMs with excellent generative capacity, such as GPT-4 \citep{gpt-4} have been presented. While it is tempting to focus our research on them, we focus on small to medium size LMs in a fine-tuning setup. 
This choice is because utilizing a huge LM as a teacher is often infeasible, e.g., due to their high financial costs or when the data cannot be sent to external servers because of privacy constraints.
Furthermore, research suggests that using mediator-teachers aids the distillation process \citep{teacher_assist}, as might be the case in distillation from a huge LM to a medium fine-tuned teacher and finally to a small student. 
For an extended discussion, see \S\ref{sec:limitations}.

Our work hence focuses on a medium size fine-tuned teacher and we assume there are several thousand labeled examples for its fine-tuning. Despite the above limitations, applying huge LMs in some valuable setups is still possible. Therefore, we also consider the distillation of one such model (GPT-4), although this is not our main focus.

We start our study by comparing architectural (Encoder-decoder vs. Decoder-only), pruning and KD design decisions, discussing the tradeoff between computational resources and task performance. We focus on practical measures like \emph{latency} and  \emph{throughput}, which is important for batch-offline applications and is typically overlooked.

We next provide the first exposure bias perspective for KD which motivates PT augmentation. This bias derives from teacher forcing when the LM conditions on ground-truth tokens during training, while at inference time it conditions on previously generated tokens \citep{rnn_exposure}. As the distillation progresses, the student's predictions gradually become similar to its teacher's, and therefore training with PTs can alleviate exposure bias. 

We propose extensions of the common practice of generating a single mode approximation PT via beam search, instead, we suggest sampling multiple PTs to facilitate higher exposure to conditional distribution factors. Additionally, we generate PTs for unlabeled data and demonstrate their effectiveness. Moreover, we propose a novel KD technique termed \emph{Joint-Teaching}, which applies word-level KD to PTs generated by both the teacher and the student. This technique aims to implicitly and explicitly address the student exposure bias, ground the learning and teach it to correct its mistakes.

Finally, we extend the scope of our study by working solely with limited unlabeled examples. Due to the absence of labeled examples, fine-tuning the teacher is infeasible, leading us to depend on huge LMs with zero-shot capabilities. Consequently, we investigate whether our KD findings from our realistic setup (which involves a fine-tuned teacher) remain applicable to the new extreme setup. To this end, we show how to successfully distill GPT-4, a huge Decoder-only model, into a small Encoder-decoder model (T5-small), which also has a different tokenizer. 

Our main empirical findings (\S\ref{sec:results}) are: (1) Encoder-decoder architectures outperform their Decoder-only counterparts in task-specific fine-tuning for NLG; (2) Decoder pruning substantially outperforms encoder pruning when considering both latency and task performance; and (3) PTs can be used much more effectively compared to what was suggested in previous work and this yields substantially improved task performance on a much reduced computational cost.

\section{Background and Related Work}
\label{sec:background}

\subsection{Natural Language Generation}
\label{sub:nlg}

Modern LMs based on Transformers leverage two primary architectures for text generation: \emph{Encoder-decoder (ED)} \citep{attention_is_all} and \emph{Decoder-only (DO)} \citep{gpt2}. While ED models are more popular for classification, summarization, and NMT, DO models excel on open-text generation and zero/few-shot setups \citep{zero_shot_arch}. 

Nonetheless, the increasing popularity of massive DO models like GPT-3 and PaLM \citep{gpt3, palm} with impressive generation capabilities, has led to the question of \emph{whether ED models are still relevant for NLG?} In \S\ref{sec:results} and in the appendix (\S\ref{sec:lms}) we discuss and demonstrate the differences between these two architectures. We show in line with recent work of \citet{ul2} that ED models outperform DO models in task-specific fine-tuning for conditional NLG. In light of this observation, we focus on KD only for ED models in the rest of the paper.

Text generation is a structured prediction problem where the goal is to sample a text $\hat{y}$ from the distribution learned by the LM, conditioned on the input text $x$. The training objective of the LM is to minimize the \emph{negative log-likelihood (NLL)} of the training dataset, by factorizing $-\log{P(y|x)}$ into $-\sum_{i=1}^{|y|}{\log{P(y_i|x,y_{<i})}}$. At inference time, the LM generates one token at a time according to the conditional distribution: $P(y_i|x,\hat{y}_{<i})$. The selection of the next token is handled by the decoding method. Beam search, which aims to find the most likely target, is the the de-facto standard \citep{decoding_survey}. Alternatively, it is possible to frame decoding as sampling, as we do in this work.

\linespread{0.9}

\subsection{Exposure Bias}
\label{sub:exposure}

LMs learn the distribution $P(y|x,y_{<i})$ at the training phase by conditioning on the ground truth $y_{<i}$. This is known as \emph{teacher forcing} which makes the training efficient and stable but also creates a mismatch at inference time, since the LM conditions on its previous predictions $\hat{y}_{<i}$. This discrepancy between training and inference is called \emph{exposure bias}. Potential side-effect is that a single error during generation may have a cascading effect by causing a deviation from the ground truth distribution and resulting in an accumulation of errors \citep{exposure_bias_imitating}. Many works link exposure bias to generalization, hallucinations, and degeneration \citep{exposure_bias_closer_look, exposure_bias_degration}.

Recent works attempted to address exposure bias, most of which focused on open-text generation and NMT \citep{exposure_bias_closer_look, on_exposure_bias_nmt, exposure_bias_oracles}. Other works addressed this problem by applying reinforcement learning techniques \citep{rnn_exposure} or by \emph{scheduled sampling} which replace ground truth tokens with generated tokens \citep{exposure_bias_scheduling, exposure_bias_scheduling2}. However, it leads to training  with inaccurate and noisy signals \citep{exposure_limitation}. In contrast to other works which study this problem in a general setting, in KD setting the teacher can be used to mitigate the student exposure bias by utilizing PTs and reliable signals from it. This is the first work to discuss exposure bias in KD.

\subsection{Compression and Knowledge Distillation}
\label{sub:kd}

There has been extensive research on model compression on techniques such as parameter sharing, pruning, quantization and factorization \citep{kd_cv_survey, compression_survey, efficient_survey, kd_survey_xu}. \emph{Pruning} \citep{first_pruning} aims to discard unimportant weights of a pre-trained or fine-tuned LM, making it more efficient while preserving performance. Usually, the pruning is structured, and complete blocks, rows, or layers are removed according to their magnitude, changes during training \citep{movement_pruning}, or causal effect \citep{causal_pruning}. 

Typically there is a performance gap between the original and  the compressed model, which can be closed by applying \emph{Knowledge distillation (KD)} \citep{first_kd} -- a technique for transfering knowledge from a large trained model (teacher $T$) to a smaller one (student $S$), by training the student to mimic the teacher's predictions or features. KD can be divided into two categories: \emph{task-agnostic}, where the goal is to mimic a pre-trained LM's behavior, and \emph{task-specific}, where the distillation is performed on a fine-tuned LM. Generally, there are three levels of KD: word-level (or class-level), inner-level, and sequence-level (only in NLG): 

\textbf{Word-level KD}, also known as \emph{Logits KD} \citep{first_kd, sequence_level}. In this method, the student learns to match the teacher's distribution over the next token at each position, by minimizing the KL divergence between the distribution of the student $P_{S}(y_{i}|x,y_{<i})$ and the distribution of its teacher $P_{T}(y_{i}|x,y_{<i})$. 
There are variations like \emph{Noisy KD} \citep{noisy_self} where noise is injected during KD by applying dropout to the teacher, \citet{selective_kd} which applies KD only for carefully selected examples, etc. 

\textbf{Inner-level KD} aims to mimic additional inner features of the teacher, for example, \citet{tinybert} leverages hidden states of the teacher to train the student.
\citet{minilm} and \citet{minilm_v2} proposed \emph{Attention-relations KD} which 
trains the student to mimic the relation matrix (scaled dot-product) of the self-attention states. 

\textbf{Sequence-level KD} is commonly used for NMT \citep{sequence_level, fast_nmt, deep_shallow}. In this approach, the teacher generates PTs for inputs in the original dataset, and student is trained to predict them. Usually, the teacher generates a single PT using beam search, which is known as ``mode approximation'' of $P_{T}(y|x)$. 

\begin{figure*}[t]
    \centering
    \includegraphics[width=\textwidth]{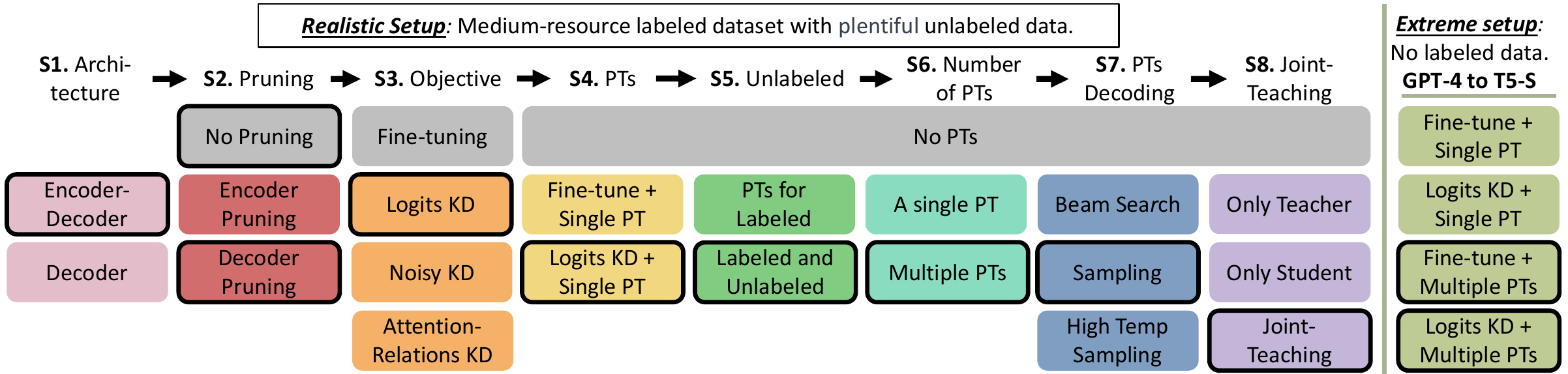}
    \caption{The design of our research. At each stage (from left to right), we examine different modeling decisions in order to gain a better understanding of their impact. We start with the architectural decisions which largely impact the task performance and computational aspects of the NLG models. Following that, we compare different KD objectives, and then we focus on augmenting the training data with Pseudo-Targets (PTs). A bold border indicates the decision we made at each stage based on the average development set performance over four NLG tasks.}
    \label{fig:diagram}
    \vspace{-1em}
\end{figure*}

\citet{kd_cv_survey} and \citet{compression_survey} present a detailed overview of KD techniques. Notably, most works in NLP explore task-agnostic KD for encoder-only models \citep{distilbert, tinybert, minilm_v2} or focus on NMT \citep{sequence_level, deep_shallow, selective_kd}. \citet{kd_summarization} focused on high-resource summarization, and compared three KD strategies: pruning and fine-tuning, logits KD, and mode approximation PTs. 
Unlike these works, we perform a more systematic study of task-specific KD for a variety of NLG tasks in realistic setups. Moreover, we focus on PTs and propose extensions to demonstrate their effectiveness.

\section{Methods}
\label{sec:methods}

\linespread{0.6}
\subsection{Research Design}
\label{sub:design}

Our research design illustrated in  Figure~\ref{fig:diagram} has eight stages. At each stage, we examine different modeling decisions and continue to the next stage after selecting the best technique according to the performance on the development set (to avoid performing selection on the test set). We linearly examine one aspect at a time since the alternative (combinatorial choices)
is too expensive. Our study starts with architectural designs (\textbf{stages 1-2}), continues with comparing different KD strategies (\textbf{stages 3-4)} and proceeds to explore the usage of PTs as augmentation strategies for KD (\textbf{stages 5-8)}. 

\linespread{0.6}
\subsection{Architectures and Pruning}
\label{sub:architecture}

In the spirit of our realistic setup, we consider off-the-shelf LMs and experiment with two model families for each architecture type (see \S\ref{sub:models}). In appendix \S\ref{sec:lms} we discuss the differences between ED (Encoder-Decoder) and DO (Decoder-only) architectures (\textbf{stage 1}) and show that ED models outperform DO models on task-specific tuning for NLG. Following that, we present results only for ED in \S\ref{sec:results}. In 
\textbf{stage 2}, we examine the effect of pruning, by discarding complete model layers. In the case of ED, layers can be dropped either from the encoder or decoder components, resulting in different impacts on the task or computational performances. 


\linespread{0.6}
\subsection{Objectives\footnote{More formal descriptions and implementation details of the methods discussed in \S\ref{sub:objective} and \S\ref{sub:pt} are provided in \S\ref{sec:formal_methods}}}
\label{sub:objective}

As discussed in \S\ref{sub:kd}, various works proposed different training strategies for KD. In \textbf{stage 3} we perform a comparison between three popular KD objectives (baselines), which do not involve PTs: (1) \emph{Logits KD} -- which is the most common and the simplest technique;
(2) \emph{Noisy KD} -- which showed promising results for summarization in self-distillation setup; and (3) \emph{Attention-Relations KD} (combined with Logits KD) -- which is the SOTA technique for Encoder-only models.

As suggested by \citet{xtremedistil}, following the end of the KD stage, we also perform an end-to-end fine-tuning stage on the ground truth labels. This stage is extremely cheap since a teacher is not required.

\begin{figure}[t]
    \centering
    \includegraphics[width=0.48\textwidth]{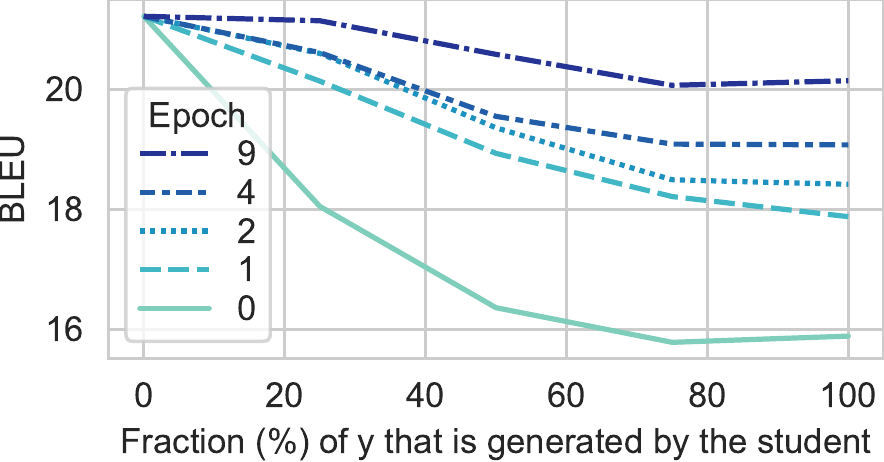}
    \caption{\bleu{} scores measured for targets generated by the \tfives{} student and the \tfivel{} teacher for the development set of the \squad{} dataset. For each input we generate X\% of the final output with the student, and then use the teacher to continue the generation and complete the output. The X-axis represents the fraction that is generated by a distilled student (at the end of the epoch). This analysis examines the effectiveness of the teacher's modeling of $P_T(y_i|x,\hat{y}_{<i}^S)$.}
    \label{fig:exposure}
    \vspace{-1em}
\end{figure}

\linespread{0.6}
\subsection{Pseudo-Targets (a.k.a sequence-level KD)\footnotemark[1]}
\label{sub:pt}

\emph{Pseudo-Targets (PTs)} are predictions generated by the teacher that can be utilized for training the student. Word-level or Inner-level KD can be combined with sequence-level KD (e.g., by applying Logits KD to PTs). In \textbf{stage 4} we investigate the impact of augmenting the labeled data with PTs when fine-tuning the student (sequence-level KD) or when using the objective from stage 3. 

Although various works demonstrated the effectiveness of PTs  \citep{sequence_level, kd_summarization}, their use of PTs was limited to a single PT per labeled training example, generated with mode approximation beam search. In this paper we demonstrate that the use of PTs can be much more extensive: We generate multiple PTs per training example, increase their diversity with sampling-based rather than mode approximation generation, and generate PTs for both labeled and unlabeled examples, which are much more abundant by nature. Our experiments demonstrate that each of these extensions yields substantial improvements in the quality of the resulting student model. We next touch on each of these extensions.

\smallskip\noindent\textbf{Unlabeled data} In our setup unlabeled data is available in abundance. Since in autoregressive NLG the LM learns to condition on the targets ($y_{<i}$), PTs are essential for utilizing unlabeled data (inputs without corresponding targets). From a generalization perspective, exposing the model to more inputs, and consequently to more $P(y_i|\boldsymbol{x},\hat{y}_{<i}^T)$ factors, should help the student generalize beyond the labeled data distribution. Indeed, many works in various NLP fields have shown that unlabeled data is effective for generalization \citep{consistency, xtremedistil, docogen}. In \textbf{stage 5} we examine its importance.

\smallskip\noindent\textbf{Multiple PTs} We further explore alternatives to the common practice of generating a single PT with beam search (mode approximation). Unlike classification, NLG is a structured prediction problem and multiple candidates can form a correct solution. Therefore, we can generate multiple PTs resulting in stronger exposure to the teacher’s knowledge. We explore the impact of multiple PTs in \textbf{stage 6}.

\smallskip\noindent\textbf{Sampling PTs} Beam search is not the only way to generate PTs. In fact, it has been demonstrated to produce generic texts that lack diversity \citep{bs_diversity2006, bs_diversity2013}. A simple alternative that can produce more diverse and surprising texts is sampling \citep{bs_diversity2020, nucleus}. Moreover, controlling the temperature of the logits can increase the diversity of the PTs even further  \citep{temperature}. We compare these decoding techniques in \textbf{stage 7}. 

\smallskip\noindent\textbf{Motivation for these Extensions} Compared to a single mode approximation PT, sampling multiple PTs for both the labeled and unlabeled examples should add more variability to the student training and cover a larger portion of the learnable distribution, which are known to improve generalization. Furthermore, these extensions expose the student to more of the teacher's knowledge.

Additionally, we also provide an exposure bias motivation. During the distillation the student's predictions gradually become similar to its teacher's predictions: $\hat{y}^S \sim \hat{y}^T$. Therefore, we can expect that training the student with diverse PTs may mitigate its exposure bias, which occurs at inference when it conditions on $\hat{y}_{<i}^S$, and not on the ground-truth distribution. In addition, PTs of unlabeled examples can help mitigate this bias as the student is getting exposed to the teacher's knowledge rather than the gold standard. Moreover, multiple and diverse PTs results in extensive exposure to additional $P(y_i|x,\boldsymbol{\hat{y}_{<i}^T})$ factors. Therefore, we hypothesize that sampling multiple PTs will improve the student compared to a mode approximation PT.

\linespread{0.6}
\subsection{Joint-Teaching}
\label{sub:co_teaching}

As mentioned above, training with PTs generated by the teacher may implicitly mitigate the student exposure bias. On the other hand, we can try to mitigate this bias explicitly by training the student while conditioning on its predictions (i.e. generate PTs with the student and use them for training). Generally, this can be unstable since the student may learn from its own mistakes. Fortunately, in KD we have a strong oracle: the teacher. By applying word-level KD on $\hat{y}_{<i}^S$, the teacher can teach the student how to continue its generated sequence correctly and prevent a cascading effect.

Nevertheless, this relies on the reasonable assumption that the teacher models $P(y_i|x,\hat{y}_{<i}^S)$ better than the student. In Figure~\ref{fig:exposure} we present a single setup analysis that supports this assumption: At almost any stage of the student training, continuing the generation with the teacher results in better predictions. Moreover, as the student becomes more similar to the teacher, we can expect the teacher to model $P(y_i|x,\hat{y}_{<i}^S)$ even better, which makes the word-level signals more reliable. This is also supported by Figure~\ref{fig:exposure}: As the distillation progresses, the teacher continuations keeps getting better.

Following that, we propose a novel KD method which addresses the exposure bias implicitly and explicitly namely \textit{Joint-Teaching}: Apply word-level KD on PTs generated by both the teacher and the student. In our experiment we randomly use the student's PTs for 50\% of the training steps. In \textbf{stage 7} we compare training only with the students’ PTs or the teachers’ PTs to Joint-Teaching, demonstrating the superiority of the latter.



\section{Experimental Setup}
\label{sec:experimental}

In this section we describe our four NLG tasks and datasets, the participating models and the evaluation procedures.  
URLs of the code and datasets, as well as implementation details and hyperparameter configurations are described in  \S\ref{sec:additional_implementation}. Additionally, a comparison between ED and DO architectures (stage 1) is provided in \S\ref{sub:transformer}; theoretical and empirical complexity analyses are provided in  \S\ref{sub:complexity}.

\linespread{0.6}
\subsection{Tasks and Datasets}
\label{sub:datasets}

We selected four English-to-English core NLG tasks, which are part of several NLG benchmarks and surveys \citep{st_survey, gem, gemv2, genie, nlg_survey, shakespeare_survey}. We built a new realistic experimental setup, in which the ratio of labeled to unlabeled data is 1:4, and the amount of labeled data is reasonable. For each task (excluding \shakespeare{}) we keep the original assignment of each example to its train-test splits. The exact numbers are provided in Table~\ref{tab:datasets}. 


\smallskip\noindent\textbf{Summarization} (\xsum{}) We use the XSUM dataset \citep{xsum} for the abstractive summarization task. The task of the NLG model is to generate an introductory sentence (summary) for a given news article. 

\smallskip\noindent\textbf{Question Generation} (\squad{}) We use the SQuAD dataset \citep{squad, squadv2} for the question generation task. Given a Wikipedia document and an answer to the question, the task of the NLG model is to generate the question. 

\smallskip\noindent\textbf{Abductive Reasoning} (\art{}) We use the $\alpha$NLG (also known as ART) dataset \citep{art} for abductive reasoning generation task. The task of the NLG model is to generate a plausible explanation for two given observations.

\smallskip\noindent\textbf{Style Transfer and Simplification} (\shakespeare{}) We construct a new dataset for the well-explored style transfer task (which is also a simplification task) of translating Shakespeare's texts to modern English. We combined pairs of Shakespearean and modern English texts from Shakespeare's plots (taken from \citet{shakespeare, shakespeare_st}), with other texts written by Shakespeare \citep{shakespeare_unsupervised} and created a  parallel style transfer dataset, see \S\ref{sub:shakespeare}.

\begin{table}
\centering
\begin{adjustbox}{width=0.5\textwidth}
\begin{tabular}{l|cccc|cc}
\toprule
 \textbf{Name} &\textbf{ Train} &\textbf{ Unlabeled} &\textbf{ Dev} &\textbf{ Test} & \textbf{Input} & \textbf{Target} \\
\midrule
 \xsum{} & 40K & 164K & 5K & 11.3K & 480 & 32 \\
 \squad{} & 17.5K & 70K & 1.57K & 9K & 320 & 32 \\
 \art{} & 10K & 40K & 1.25K & 14.3K & 48 & 32 \\
 \shakespeare{} & 7K & 28K & 0.75K & 0.8K & 48 & 48 \\
\bottomrule
\end{tabular}
\end{adjustbox}
\caption{Datasets used in our experiments and the number of examples in each split. Input and Target represent the maximum lengths (tokens) we use in experiments.}
\label{tab:datasets}
\vspace{-1em}
\end{table}

\subsection{Models and Pruning}
\label{sub:models}

\smallskip\noindent\textbf{Decoder-only} We use the GPT2-family models \citep{gpt2}: \gpts , \gptm , and \gptl ; and the recent OPT-family models \citep{opt}: \opts and \optm . 

\smallskip\noindent\textbf{Encoder-decoder} We use the T5-family models \citep{t5}: \tfives{} and \tfivel ; and the BART-family models \citep{bart}: \bartb{} (base version) and \bartl .

\smallskip\noindent\textbf{Pruning} We apply pruning only for the pre-trained \bartb{} model (thus our study also includes a non-pruned student, \tfives{}), and consider two types of pruning: Encoder pruning and decoder pruning. Following \citet{kd_summarization}, in both pruning types we keep only the first and last layers, resulting in two models: \bartp{} (pruned encoder) and \barts{} (pruned decoder). 

In the KD stages (3-8) we use two student-teacher pairs: \tfives{} and \tfivel{}, and a pair with a pruned student: \bartp{} and \bartl{}.

\subsection{Evaluation}
\label{sub:evaluation}

\smallskip\noindent\textbf{Task Performance} We report on various metrics that focus on different aspects, resulting in a more holistic evaluation of the models. To this end, we focus on the lexical similarity metrics, \bleu{} and \rouge{}, the semantic equivalence metric BERTScore (\bertscore{}, \citet{bertscore}) and the statistical modeling metric Perplexity (PPL), which is measured by the average NLL of the ground truth targets. To make the result tables more readable, we report the average \rouge{} (of the F1 scores for R-1/2/L), and the F1 score for \bertscore{}. Notice that in \S\ref{sec:additional_implementation} we specify for each task the appropriate metric we use for the development set. In \S\ref{sec:additional_results} we report the scores of all the metrics.

\def\arraystretch{0.85}
\begin{table*}
\centering
\begin{adjustbox}{width=\textwidth}
\begin{tabular}{ll|ccc|ccc|cccc|c}
\toprule
\textbf{Arch} & \textbf{Model} & \textbf{E-D} & \textbf{Params} & \textbf{Mem} & \textbf{FLOPs} & \textbf{Latency} & \textbf{Throughput} & \textbf{BLEU} & \textbf{ROUGE} & \textbf{BS} & \textbf{PPL} & \textbf{Dev} \\
\midrule
 DO & \gptl & 0-36 & 774 & 3210 & 42.0 & 675 & 2.2K & 11.9 & 27.1 & 70.1 & 1.9 & 13.0 \\
 DO & \gptm & 0-24 & 354 & 1444 & 19.4 & 459 & 4.8K & 9.7 & 23.2 & 66.8 & 3.7 & 10.8 \\
 DO & \gpts & 0-12 & 124 & 511 & 6.8 & 235 & 13.5K & 7.8 & 20.1 & 61.4 & 2.8 & 8.5 \\
 DO & \optm & 0-24 & 331 & 1324 & 18.1 & 371 & 5.1K & 9.8 & 24.9 & 62.7 & 3.1 & 10.7 \\
 DO & \opts & 0-12 & 125 & 502 & 6.8 & 185 & 15.4K & 10.7 & 26.3 & 69.2 & 2.5 & 11.7 \\
\midrule
 ED & \underline{\tfivel} & 24-24 & 737 & 2951 & 19.5 & 597 & 5.3K & 16.4 & 34.6 & 75.1 & 1.6 & 17.7 \\
 ED & \textbf{\tfives} & 6-6 & 60 & 242 & 1.4 & 160 & 55.2K & 13.4 & 30.8 & 72.7 & 2.4 & 14.6 \\
\midrule
 ED & \underline{\bartl} & 12-12 & 406 & 1625 & 10.0 & 281 & 7.8K & 16.4 & 34.8 & 75.4 & 1.7 & 17.9 \\
 ED & \bartb & 6-6 & 139 & 558 & 3.0 & 147 & 13.5K & 14.5 & 32.7 & 74.2 & 1.9 & 15.9 \\
 ED & \bartp & 2-6 & 111 & 445 & 1.7 & 146 & 16.0K & 11.4 & 28.0 & 71.6 & 2.2 & 12.8 \\
 ED & \textbf{\barts} & 6-2 & 101 & 407 & 2.6 & 75 & 15.3K & 13.3 & 31.5 & 73.3 & 2.6 & 15.0 \\
\bottomrule
\end{tabular}
\end{adjustbox}
\caption{A comparison between different architectures and models, fine-tuned on our datasets. E-D represents the number of Encoder and Decoder layers. All the computational and performance measures are the average over four NLG tasks. Params are in millions, Mem is in MB, FLOPS is in billions, Latency is in milliseconds. Throughput is in thousands of examples per minute. Everything is measured on an Nvidia GeForce RTX 4080. We also indicate the participating models in the KD stages (3-8): Teachers are underlined, student baselines (no KD) are in bold.
}
\label{tab:performance}
\end{table*}

\smallskip\noindent\textbf{Computational Performance} For measuring the computational performance of the models, we report the number of parameters, the memory of the models and the number of \emph{floating-point operations (FLOPs)}. These measures are device-agnostic and may not be well correlated with the actual performance in practice, which depends on the device, implementation, and hardware utilization of the accelerators \citep{shufflenet, throughput}. Therefore, we also report practical measurements such as the \emph{latency} of generating a single output, which is important for real-time applications, and the \emph{throughput}, which is the maximum number of examples that can be processed in a minute, and is important for offline batched applications. 





\section{Results}
\label{sec:results}

The complete results are provided in \S\ref{sec:additional_results}.
Table~\ref{tab:performance} reports the results of fine-tuned models (stages 1-2).
Table~\ref{tab:kd} reports the results of the KD stages (3-8) as follows: For each student-teacher pair and dataset, we calculate the fraction of their performance gap that is compensated for by using distillation as opposed to only fine-tuning the student model: $\frac{KD-S}{T-S}\%$, where $KD$, $T$ and $S$ are the task scores of the distilled student, its teacher and the student baseline (fine-tuned), respectively. Then, we report for each dataset the average fraction of the closed gap over four metrics and two student-teacher pairs. We also report the number of wins within 32 setups (4 datasets, 4 metrics, 2 pairs).

\medskip\noindent\textbf{S1: Encoder-decoder models outperform Decoder-only models in task-specific tuning for NLG.} 
 We present our results in Table~\ref{tab:performance}. For a detailed analysis of Encoder-decoder (ED) and Decoder-only (DO) models, we refer readers to Appendix \S\ref{sec:lms}, which reports several interesting theoretical and empirical insights.
Nevertheless, it is worth noting here that ED models, such as \tfivel{}, can have twice the number of layers and parameters of DO models, such as \gptm{} or \optm{}. However, despite the higher number of parameters, ED models have roughly the same FLOPs and comparable latency and throughput.

Regarding task performance, our experiments demonstrate that ED models consistently outperform DO models across all datasets and models, regardless of their size. Presumably, a better inductive bias is injected by applying self-attention (and not autoregressive-attention) to the conditioned input sequence. 
This finding is particularly relevant for NLP practitioners who aim to develop a specialized in-house model for a specific NLG task. We hence continue to the KD stages only with ED models (T5 and BART). 

\medskip\noindent\textbf{S2: It is better to prune layers from the decoder.} 
In stage 2, we examine whether it is better to prune encoder or decoder layers. To this end, we prune \bartb{} and report the results at the bottom of Table~\ref{tab:performance}. First, notice that pruning decoder layers greatly impacts the latency given the autoregressive nature of NLG tasks, making \barts{} two times faster than \bartb{}. For comparison, pruning encoder layers does not affect the latency (see the discussion in \S\ref{sub:complexity}). On the other hand, \bartp{} has a higher throughput than \barts{}, mainly because of the long input in some tasks which is processed by the encoder. Notice, however, that the improvement of \barts{} in latency is more substantial than its throughput degradation.

Second, \barts{} outperforms \bartp{} in every task metric (and dataset), being competitive to \bartb{}. Moreover, for tasks with long inputs (e.g., summarization or question generation, see \S\ref{sec:additional_results}), the depth of the encoder is critical and the pruned-encoder \bartp{} underpeforms. As a rule of thumb, our results suggest that it is better to prune layers of the decoder. Besides reducing the model latency, it has a smaller impact on task performance. In the following stages we use two student-teacher pairs: \tfives{} and \tfivel{}, and a pair with a pruned student, \barts{} and \bartl{}. 

\medskip\noindent\textbf{S3: Use Logits KD as the main training objective.}
In stage 3 we compare different KD objectives. As seen in Table~\ref{tab:kd}.A, Logits, Noisy and Attention-Relations KD techniques are competitive, and the quality of the method depends on the task. Even though Noisy KD has more wins than Logits KD, the \ppl{} metric accounts for 8 of the 14 wins. Since Logits KD is the best-performing method according to the average performance on the development set, we continue to the next PT stages with it. Our results demonstrate the importance of KD: applying Logits KD closes more than 34.4\% of the student-teacher gap, on average. 

\def\arraystretch{0.8}
\begin{table}
\centering
\begin{adjustbox}{width=0.5\textwidth}
\begin{tabular}{l|cccc|c|c}

\toprule
A. \textbf{Objective} & \textbf{XS} & \textbf{SQ} & \textbf{AR} & \textbf{SH} & \textbf{Wins} & \textbf{Dev} \\
  & (\%) & (\%) & (\%) & (\%) & &\\
\midrule
 Fine-tune & 0.0 & 0.0 & 0.0 & 0.0 & 0 & 14.8 \\
 Logits & 30.2 & \textbf{39.7} & 25.7 & \textbf{41.9} & 13 & \textbf{16.0} \\
 Noisy & 30.3 & 37.3 & \textbf{35.2} & 41.8 & \textbf{14} & 15.9 \\
 Att-Rel & \textbf{31.3} & 28.4 & 19.7 & 21.4 & 5 & 15.9 \\

\midrule
 B. \textbf{PTs} & \textbf{XS} & \textbf{SQ} & \textbf{AR} & \textbf{SH} & \textbf{Wins} & \textbf{Dev} \\
\midrule
Logits & 30.2 & \textbf{39.7} & 25.7 & 41.9 & 10 & 16.0 \\
Seq-lvl & 13.8 & -9.1 & 4.2 & 4.2 & 0 & 15.7 \\
Logits+Seq & \textbf{33.2} & 30.8 & \textbf{27.9} & \textbf{49.0} & \textbf{22} & \textbf{16.3} \\

\midrule
 C. \textbf{Unlabeled} & \textbf{XS} & \textbf{SQ} & \textbf{AR} & \textbf{SH} & \textbf{Wins} & \textbf{Dev} \\
\midrule
Labeled & 33.2 & 30.8 & 27.9 & 49.0 & 0 & 16.3 \\
 + Unlabeled & \textbf{55.8} & \textbf{47.1} & \textbf{41.5} & \textbf{70.0} & \textbf{32} & \textbf{16.9} \\

\midrule
 D. \textbf{Decoding} & \textbf{XS} & \textbf{SQ} & \textbf{AR} & \textbf{SH} & \textbf{Wins} & \textbf{Dev} \\
\midrule
 Single PT & 55.8 & 47.1 & 41.5 & 70.0 & 1 & 16.9 \\
 K-Beams & 63.6 & 56.3 & 45.7 & 74.7 & 4 & 17.0 \\
 Sampling & \textbf{73.0} & 58.4 & \textbf{48.2} & 81.7 & \textbf{15} & \textbf{17.2} \\
 H-Sampling & 70.0 & \textbf{63.9} & 44.8 & \textbf{81.8} & 12 & 17.1 \\

\midrule
 E. \textbf{Joint-T} & \textbf{XS} & \textbf{SQ} & \textbf{AR} & \textbf{SH} & \textbf{Wins} & \textbf{Dev} \\
\midrule
 Only Teacher & 73.0 & 58.4 & \textbf{48.2} & 81.7 & 4 & 17.2 \\
 Only Student & 68.7 & 63.9 & 43.9 & 79.4 & 3 & 17.1 \\
 Joint-Teaching & \textbf{80.8} & \textbf{66.7} & \textbf{48.2} & \textbf{87.7} & \textbf{25} & \textbf{17.4} \\

\bottomrule
\end{tabular}
\end{adjustbox}
\caption{Average fractions of the student-teacher {\em performance gaps closed} by different KD methods, number of winning setups (out of 32) and development scores. Table A: KD objectives (S3); Table B: PT augmentation (S4). Table C: PTs for unlabeled examples (S5); Table D: Decoding methods for generating PTs (S6+S7); Table E: PTs generated only by the teacher, the student, or Joint-Teaching (S8). For more details about the methods see \S\ref{sec:methods} and \S\ref{sec:formal_methods}. Tasks: \xsum{} (XS), \squad{} (SQ), \art{} (AR), \shakespeare{} (SH).}
\label{tab:kd}
\vspace{-1em}
\end{table}

\medskip\noindent\textbf{S4: Combine Logits KD and PTs.}
In stage 4 we examine three methods: using Logits KD only on the labeled examples, fine-tuning the student with PTs (Sequence-level KD) or combining them. The corresponding rows in Table~\ref{tab:kd} show that sequence-level KD underperforms Logits KD. However, their combination results in a better student in 22 setups and achieves a higher development score, and therefore, we use this strategy in the subsequent stages.

\medskip\noindent\textbf{S5: Unlabeled data should be utilized.}
Generating PTs for the unlabeled inputs may help extract more of the knowledge embodied in the teacher, allowing the student to generalize better. In stage 5 we explore this hypothesis.
According to Table~\ref{tab:kd}.C, utilizing unlabeled data greatly boosts the performance and closes an additional 19\% of the gap. 
To the best of our knowledge, this is the first study that shows this in KD for NLG. 
In the next stages, we generate PTs for the labeled and unlabeled inputs.

\medskip\noindent\textbf{S6: Exposing the student to multiple PTs helps.} 
By comparing the rows of Single PT and K-Beams in Table~\ref{tab:kd}.D, it can be seen that exposing the student to multiple targets and covering a larger portion of learnable distribution closes an additional 6.4\% of the gap on average. 

\medskip\noindent\textbf{S7: Sampling is better than Beam-Search for generating PTs.}
Table~\ref{tab:kd}.D also shows that generating PTs with sampling is typically better than beam search, and closes another 5.2\% of the gap on average. We observe that high sampling temperature is competitive, although its effect depends on the task and model. High sampling works better for \tfives{}, while sampling without temperature works better for \barts{} (and on average). Further research could investigate a larger range of temperatures and other diversity-oriented decoding methods.
Nevertheless, this is the first study that challenges the traditional mode-approximation practice, and show that generating multiple PTs via sampling significantly improves NLG distillation.

\medskip\noindent\textbf{S8: Joint-Teaching improves the student.}
The results in Table~\ref{tab:kd}.E support two of our hypotheses, which we discuss in \S\ref{sub:co_teaching}. The first is that PTs generated only by the student are less valuable for its training than PTs generated by teacher. The second is that the combination of the two types of PTs (by Joint-Teaching) can be more effective for KD than using only PTs generated by the student or teacher. Our Joint-teaching approach wins 25 out of 32 times and closes another 5.7\% of the gap.


\def\arraystretch{0.85}
\begin{table*}[htbp]
\centering
\begin{adjustbox}{width=\textwidth}
\begin{tabular}{l|l|ccc|cccc}
\toprule
 \textbf{Dataset} & \textbf{Model} & \textbf{FLOPs} & \textbf{Latency} & \textbf{Throughput} & \textbf{BLEU} & \textbf{ROUGE} & \textbf{BScore} & \textbf{PPL} \\
\midrule
 \multirowcell{4}{XSUM\\40K} & \tfivel & 38.7 & 539 & 1.3K & 11.5 & 29.3 & 72.7 & 1.7 \\
 & \tfivekd & \multirowcell{1}{2.7 (-93\%)} & \multirowcell{1}{144 (x3.7)} & \multirowcell{1}{13.4K (x10.3)} & 10.7 (80\%) & 28.2 (81\%) & 71.8 (80\%) & 1.9 (87\%) \\\cmidrule{2-9}
 & \bartl & 19.6 & 254 & 3.3K & 13.0 & 31.1 & 73.9 & 1.7 \\
 & \bartkd & \multirowcell{1}{5.1 (-73\%)} & \multirowcell{1}{68 (x3.7)} & \multirowcell{1}{10.0K (x3.0)} & 12.3 (79\%) & 30.2 (79\%) & 73.5 (84\%) & 1.9 (73\%) \\\cmidrule{1-9}
 \multirowcell{4}{SQuAD\\17.5K} & \tfivel & 26.1 & 530 & 2.0K & 22.2 & 42.3 & 77.9 & 1.3 \\
 & \tfivekd & \multirowcell{1}{1.8 (-93\%)} & \multirowcell{1}{143 (x3.7)} & \multirowcell{1}{22.3K (x11.1)} & 20.9 (57\%) & 40.6 (57\%) & 77.0 (50\%) & 1.5 (57\%) \\\cmidrule{2-9}
 & \bartl & 13.3 & 250 & 4.8K & 21.5 & 41.9 & 77.8 & 1.4 \\
 & \bartkd & \multirowcell{1}{3.4 (-74\%)} & \multirowcell{1}{67 (x3.7)} & \multirowcell{1}{13.0K (x2.7)} & 20.9 (84\%) & 40.9 (75\%) & 77.3 (77\%) & 1.7 (71\%) \\\cmidrule{1-9}
 \multirowcell{4}{ART\\10K} & \tfivel & 5.9 & 533 & 10.7K & 6.0 & 21.7 & 71.5 & 1.9 \\
 & \tfivekd & \multirowcell{1}{0.5 (-92\%)} & \multirowcell{1}{142 (x3.7)} & \multirowcell{1}{109.8K (x10.3)} & 4.8 (49\%) & 19.9 (50\%) & 70.4 (47\%) & 2.4 (25\%) \\\cmidrule{2-9}
 & \bartl & 3.2 & 250 & 13.7K & 6.0 & 21.4 & 71.5 & 2.1 \\
 & \bartkd & \multirowcell{1}{0.8 (-75\%)} & \multirowcell{1}{67 (x3.7)} & \multirowcell{1}{23.4K (x1.7)} & 5.1 (59\%) & 20.3 (57\%) & 71.0 (61\%) & 2.4 (34\%) \\\cmidrule{1-9}
 \multirowcell{4}{Shakespeare\\7K} & \tfivel & 7.2 & 789 & 7.4K & 25.7 & 45.4 & 78.4 & 1.5 \\
 & \tfivekd & \multirowcell{1}{0.6 (-91\%)} & \multirowcell{1}{212 (x3.7)} & \multirowcell{1}{75.3K (x10.1)} & 25.7 (100\%) & 45.3 (98\%) & 78.1 (79\%) & 1.7 (56\%) \\\cmidrule{2-9}
 & \bartl & 3.9 & 367 & 9.2K & 25.1 & 44.8 & 78.3 & 1.8 \\
 & \bartkd & \multirowcell{1}{1.0 (-75\%)} & \multirowcell{1}{96 (x3.8)} & \multirowcell{1}{14.8K (x1.6)} & 24.8 (88\%) & 45.2 (123\%) & 78.1 (86\%) & 2.0 (68\%) \\
\bottomrule
\end{tabular}
\end{adjustbox}
\caption{\tfivekd{} and \bartkd{} are the final distilled models trained with Joint-Teaching. The numbers in parentheses represent computational improvements or the fraction of the {\em student-teacher gap closed} by the distilled model.}
\label{tab:final}
\vspace{-1em}
\end{table*}

\medskip\noindent\textbf{Final Compression Results.} The final compression results (after stage 8) are provided in Table~\ref{tab:final}. We attempt to achieve high compression rates: \tfivekd{} and \bartkd{} reduce 92\% and 75\% of their teachers' parameters, respectively. This results in great computational performance improvements. Our distilled models reduce the latency of their teachers by a factor of 3.7. In addition, \tfivekd{} has a 10 times higher throughput, and \bartkd{} has double the throughput of its teacher. Our study shows that KD allows model compression and drastically improves the task performance compared to the fine-tuned baseline. In most setups, our recipe for KD closes more than 75\% of the student-teacher gap. 
Surprisingly, in some of the tasks like \shakespeare{} the distilled model outperforms its teacher.
Finally, we also conduct a human evaluation to examine the relatively lower performance of our KD method on the \art{} dataset (see appendix \S\ref{sec:human_art}). Our human evaluation results show that the distilled model (\tfivekd) closes 72\% of the gap, and this is in-line with the performance on other datasets.



\subsection{Extreme setup: KD with GPT-4}
\label{sub:extreme}

In the final phase, we explore the transferability of our KD conclusions to an \emph{extreme setup} which involves only limited unlabeled examples. As labeled examples are unavailable, fine-tuning the teacher becomes impractical, leading to the reliance on a huge LM with zero-shot capabilities as the teacher, and this poses new challenges: (1) The teacher is a huge Decoder-only model (since this is the standard for zero-shot learning) while the student is an Encoder-decoder model; (2) The teacher and the student have different tokenizers and (3) Querying the teacher is financially costly, limiting its usage. 

We utilize GPT-4 \citep{gpt-4} as our teacher and \tfives{} as the student. The prompt of GPT-4 consists of three labeled demonstrations. Due to its high cost, we conduct experiments only for the \squad{} (3000 examples) and the \shakespeare{} (1500 examples) datasets, and with the following baselines and methods: (a) The GPT-4 teacher; (b) \tfives{} training with ground-truth (GT) labels; (c) Student fine-tuning with a single PT; (d) Fine-tuning with multiple (five) PTs; (e) Student training with Logits KD and a single PT (f) Logits KD with multiple PTs;
More details are provided in \S\ref{sec:extreme}.

Our results in Table~\ref{tab:kd_gpt_full} (appendix \S\ref{sub:extreme_results}) are mixed: Generating multiple PTs outperforms a single PT, but Logits KD only helps in the \squad{} dataset. Future research is called for as we attribute this result to challenges in aligning the tokenizers.

\linespread{0.6}
\section{Conclusion}
\label{sec:conclusion}

In this paper, we present a general KD recipe for NLG. To this end, we conduct a systematic study on various tasks and evaluate the impact of different 
modeling decisions on computational and task performance of distilled models. Our results suggest that using ED models as students, pruning decoder layers, combining Logits KD and PTs via sampling and Joint-Teaching achieve high compression rates while maintaining competitive performance. 

Nevertheless, our recipe is based on average performance and may depend on the task, model, or setup. The teacher-student performance gap that still exists demonstrate the need for further research. For example, high-temperature PTs seem to be less effective for BART, and further exploration of different hyperparameters or methods for increasing PT diversity may be necessary. Integrating a smart selection of training examples or PTs \citep{selective_kd}, refining Joint-Teaching with curriculum learning or scheduling \citep{exposure_bias_scheduling2} are some future research directions. 

\section{Limitations}
\label{sec:limitations}

\noindent\textbf{Using a medium size fine-tuned teacher.} 

With recent advances in huge LM such as GPT-4 and their extraordinary generation capabilities, one may wonder about the relevance of this work which mainly focuses on a medium size fine-tuned teacher. Although we show the distillation of a huge LM (GPT-4), it is often infeasible.

First, when the data cannot be sent to external servers because of privacy constraints or when the domain is unique or specific (e.g., in national security settings or human conversations), huge LMs that cannot be fine-tuned may be less effective. 

Second, we have distinguished between two types of costs: computational and financial. While training a student model with a medium-size fine-tuned teacher may take a few days, the entire process is feasible since training time is typically not a limited resource. 
In contrast, generating PTs with a huge LM like GPT-4 can easily cost (many) dozens of thousands of dollars. This financial cost is often prohibitive, particularly when training a general high-quality student or several domain-specific ones. While it is possible to utilize a huge LM to obtain a limited number of labeled examples, relying on it for generating PTs for abundant unlabeled data is not feasible. Therefore, a medium size teacher is needed. 

Furthermore, research suggests that using mediator/assistant teachers aids the distillation process \citep{teacher_assist, minilm}, as might be the case in distillation from a huge LM to a medium size fine-tuned teacher, and finally to a small student. Considering the aforementioned reasons, our study holds significant relevance as it emphasizes the importance of the distillation process with a medium size teacher, regardless of whether the data is generated manually or by a huge LM.

\medskip\noindent\textbf{The scope of our realistic setup.}
While our results demonstrate the effectiveness of KD for various English-to-English NLG tasks, for the tasks that were part of the study, the output length is relatively short compared to the input (e.g., Summarization and Question Generation) or has a similar length (Abductive Reasoning, Style Transfer and Simplification). The results may differ for tasks with much longer output lengths or for non-English-to-English tasks such as NMT, data-to-text (e.g., table-to-text), multilingual, or multi-modality tasks. 

In addition, the results are applicable to our realistic task-specific setups, and some findings may vary in high-resource scenarios or when unlabeled data is unavailable. Although these scenarios may be less  relevant to NLP application developers, they are commonly studied in academic research.

\medskip\noindent\textbf{Computational training costs.}
Another limitation of our research is that we did not consider the computational costs of the KD stages.
The training time comparison between the methods was therefore overlooked. 
This is because we assumed that one-time resource usage for training could be neglected compared to the accumulated inference cost of a deployed model.

However, it is worth noting that generating PTs with the teacher for all the training and unlabeled examples is computationally expensive (it could take one to a few days, depending on the number of unlabeled examples). 
Furthermore, Joint-Teaching can also be computationally heavier than other KD methods, as the student generates PTs during the training process (although the student is fast).

In addition, different training objectives also have different costs, with some methods being more computationally intensive than others (e.g., Attention-Relation is more costly than Logits KD). 
Finally, the distillation process can be long, and multiple epochs are required until the student converges - in some setups, we trained the student for more than a few days.

\medskip\noindent\textbf{Utilizing huge LMs.}
Certain limitations arise in our extreme setup, which involves the costly utilization of huge LMs (GPT-4) provided by external companies like OpenAI. 
First, the comparison with the Joint-Teaching method is not conducted due to the need for repeated costly querying of the teacher model to extract its logits every time a PT is generated with the student. 
Nevertheless, extracting the logits of the teacher PTs (for Logits KD) and generating multiple PTs is approximately equivalent to generating a single PT. This is because the prompt, consisting of many tokens, is processed only once, and the marginal cost of generating multiple (relatively short) PTs is low.

Another limitation arises from relying on external companies to enable logit extraction (for Logits KD) and there is no assurance that this feature will be supported. For instance, in the chat versions: ChatGPT and GPT-4, logits are not accessible. 
In this work, we rely on an internal version of GPT-4, which allows us to extract its logits. 
Fortunately, we demonstrate that even without Logits KD, achieving a strong student model is possible.

\section*{Acknowledgements}
We would like to thank the area chair, the reviewers, the members of the \emph{Microsoft MSAI} team, and the \emph{NLP@Technion} team for their valuable feedback and advice.
Roi Reichart has been partially supported by the \emph{VATAT} grant on data science. 


\bibliography{custom}
\bibliographystyle{acl_natbib}

\appendix
\section{Study Methods}
\label{sec:formal_methods}

In this section, we formally describe the objectives and methods we consider in our study and discuss in \S\ref{sec:methods}. A description of the notations is provided in Table~\ref{tab:notations}. In addition, for each method we mention the stage in which we examine it and its corresponding name  in the results Table~\ref{tab:kd}. More implementation details including hyperparameters are provided in \S\ref{sec:additional_implementation}.

\medskip\noindent\textbf{Conditional Language Modeling (fine-tuning)} \\
Stages 1 and 2. ``Fine-tune'' in Table~\ref{tab:kd}.A. 

The objective of the autoregressive LM is to minimize the Negative Log Likelihood (NLL) of the training dataset:

\begin{align*}
\mathcal{L}_{\texttt{NLL}}(x,y) & = -\log{P(y|x)} \\ 
& = -\sum_{i=1}^{|y|}{\log{P(y_i|x,y_{<i})}}
\end{align*}

Notice that in our experiments we also conduct a fine-tuning stage for 10 epochs on the labeled data after the distillation stage of the following KD methods.

\medskip\noindent\textbf{Logits KD (a.k.a Word-Level KD)} \\
Stage 3. ``Logits'' in Table~\ref{tab:kd}.A and \ref{tab:kd}.B. 

The objective of the student is to minimize the KL divergence (or the Cross-Entropy) of the next token distribution of the student and the teacher:

\begin{align*}
& \mathcal{L}_{\texttt{Log}}(x,y) = \\
& -\sum_{i=1}^{|y|}{KL(P_S(y_i|x,y_{<i}) || P_T(y_i|x,y_{<i}))}
\end{align*}

\def\arraystretch{1.5}
\begin{table}[t]
\centering
\begin{adjustbox}{width=0.5\textwidth}
\begin{tabular}{m{0.125\textwidth}|p{0.375\textwidth}}
\toprule
$x$ & Input text: A sequence of $m$ tokens, $x=(x_1,...,x_m)$. \\
$y$ & Target text: A sequence of $n$ tokens, $y=(y_1,...,y_n)$. \\
$P(y_i|x, y_{<i})$ & The next token distribution that the autoregressive LM learns via teacher forcing. \\
$\hat{y}$ & Generated text: The inference output of the LM. \\
$P(y_i|x, \hat{y}_{<i})$ & The next token distribution that is used during inference for generating $\hat{y}$. \\
$T$ & The teacher LM. \\
$S$ & The student LM ($|S| \ll |T|$). \\
$\hat{y}^T$ & A pseudo target (PT) generated by the teacher model.
\\
$\hat{y}^S$ & An output generated by the student model (student PT).
\\
$P_T(y_i|x, y_{<i})$ & The teacher's next token distribution. \\
$P_S(y_i|x, y_{<i})$ & The student's next token distribution. \\
\bottomrule
\end{tabular}
\end{adjustbox}
\caption{Notations.}
\label{tab:notations}
\end{table}

\medskip\noindent\textbf{Noisy KD} \\
Stage 3. ``Noisy'' in Table~\ref{tab:kd}.A. 

For more details see \citet{noisy_self} and \S\ref{sec:additional_implementation}. 

\medskip\noindent\textbf{Attention Relation KD} \\
Stage 3. ``Att-Rel'' in Table~\ref{tab:kd}.A. 

For more details see \citet{minilm}, \citet{minilm_v2} and \S\ref{sec:additional_implementation}. 

\medskip\noindent\textbf{Fine-tune + PTs (a.k.a. sequence-Level KD)} \\
Stage 4. ``Seq-lvl'' in Table~\ref{tab:kd}.B. 

For each labeled input $x$, we use the teacher to generate a single mode approximation PT via beam search: $\hat{y}^T$. Then we fine-tune the student by minimizing $\mathcal{L}_{\texttt{NLL}}(x,\hat{y}^T)$. 

Notice that in our experiments we actually minimize $\mathcal{L}_{\texttt{NLL}}(x,\hat{y}^T) + \mathcal{L}_{\texttt{NLL}}(x,y)$, i.e., an interpolation between the ground truth target and the PT. We find this interpolation to work better than using only the PT. \citet{sequence_level} proposed another interpolation, by selecting the most similar PT to the ground truth from a set of $K$ PTs generated by beam search. 

\medskip\noindent\textbf{Logits KD + PTs} \\
Stage 4 and 5. ``Logits+Seq'' in Table~\ref{tab:kd}.B and ``Labeled'' in Table~\ref{tab:kd}.C. 

Same as ``Fine-tune + PTs'', but we train the student to minimize: $\mathcal{L}_{\texttt{Log}}(x,\hat{y}^T)$. Following the note above, we actually minimize the interpolation: $\mathcal{L}_{\texttt{Log}}(x,\hat{y}^T) + \mathcal{L}_{\texttt{Log}}(x,y)$ (this is also the case for the following methods).

\medskip\noindent\textbf{Logits KD + PTs for unlabeled inputs} \\
Stages 5 and 6. ``+Unlabeled'' in Table~\ref{tab:kd}.C and ``Single PT'' in Table~\ref{tab:kd}.D. 

Same as ``Logits KD + PTs'', but we also generate a single mode approximation PT for each unlabeled input. 

\medskip\noindent\textbf{Logits KD + Multiple PTs} \\
Stage 6. ``K-Beams'' in Table~\ref{tab:kd}.D. 

We use the teacher to generate $K$ PTs for every labeled or unlabeled input, using beam search with a beam size of $K$. We kept all the final $K$ beams (sequences), $Y_K$, and used them to distill the student by minimizing: 
$\sum_{\hat{y}^T \in Y_K}{\mathcal{L}_{\texttt{Log}}(x,\hat{y}^T)}$.

This technique can be viewed as generating the top-$K$ mode approximations. In our experiments we use a different single PT for each input at every epoch (i.e., if we generate $K$ PTs for each input, it takes $K$ epochs until the student sees all of them). We mainly do it for a fair comparison between the different methods (see \S\ref{sec:additional_implementation} for additional details).

\medskip\noindent\textbf{Logits KD + Sampling Multiple PTs} \\
Stage 7 and 8. ``Sampling'' in Table~\ref{tab:kd}.D and ``Only Teacher'' in Table~\ref{tab:kd}.E. 

Same as ``Logits KD + Multiple PTs'', but rather than generating PTs via beam search, we sample them. Notice that in every distillation epoch a different single PT is sampled. 

\medskip\noindent\textbf{Logits KD + High Temperature Sampling of Multiple PTs} \\
Stage 7. ``H-Sampling'' in Table~\ref{tab:kd}.D. 

Same as ``Logits KD + Sampling Multiple PTs'', but we apply softmax temperature adjustment to the next token distribution when we sample PTs. High temperature values cause the next token distribution to be more flat (and increase its entropy). Therefore, high-temperature sampling generates more diverse and surprising PTs \citep{temperature}. We use $\tau=1.5$ in our experiments. 

\medskip\noindent\textbf{Logits KD + Student PTs} \\
Stage 8. ``Only Student'' in Table~\ref{tab:kd}.E. 

Same as ``Logits KD + Sampling Multiple PTs'', but instead of generating PTs with the teacher, we use the student to generate PTs. We generate PTs on-the-fly since the student is continuously updated during training. In other words, for every training input, we use the student to sample a student PT $\hat{y}^S$. Then, we calculate $\mathcal{L}_{\texttt{Log}}(x,\hat{y}^S)$ and update the student weights. The process is repeated for every input until the student finishes the training.

\medskip\noindent\textbf{Joint-Teaching} \\
Stage 8. ``Joint-Teaching'' in Table~\ref{tab:kd}.E. 

This method combines ``Logits KD + Sampling Multiple PTs'' and ``Logits KD + Student PTs''. Accordingly, we generate a PT for every training input using either the teacher or the student. The student is trained to minimize:

\vspace{-0.5cm}
\begin{align*}
\alpha\mathcal{L}_{\texttt{Log}}(x,\hat{y}^T) + (1-\alpha)\mathcal{L}_{\texttt{Log}}(x,\hat{y}^S)
\end{align*}

Where in our experiments $\alpha=0.5$, since we find it to work nicely. However, in future extensions of this method, $\alpha$ can also be a scheduled variable or a variable that depends on the student's learning.

\section{Language Models Architectures}
\label{sec:lms}

As discussed in \S\ref{sec:methods}, the first stage (\textbf{stage 1}) of our study is to select the backbone architecture of the NLG model. In this section, we thoroughly discuss and demonstrate the differences between the two common transformer architectures for NLG: Encoder-decoder (ED) models and Decoder-only (DO) models. We start by providing a background on these architectures in \S\ref{sub:transformer}. Following that, in \S\ref{sub:complexity} we present a theoretical and empirical complexity analysis. Finally, in Subsection \S\ref{sub:performance} we compare various off-the-shelf LMs from different families by fine-tuning them on several NLG tasks in the realistic setups we consider in this work.

An important note: We acknowledge that the generation capabilities of huge LMs such as GPT-3, GPT-4, and PaLM are exceptional. We do not claim that Encoder-decoder models outperform huge Decoder-only models. We consider fine-tuned small or medium-sized LMs since our teachers and students are such. In this case, Encoder-decoders are preferable for task-specific fine-tuning of NLG.

\subsection{Transformer Background}
\label{sub:transformer}

Modern LMs are based on the Multi-layer Transformer architecture \citep{attention_is_all}. The core building block of the Transformer is Attention, which processes the sequence by replacing each token with a weighted average of the rest of the sequence (self-attention), the preceding tokens (autoregressive-attention), or another input sequence (cross-attention). For text generation, there are two dominant types of models: \emph{Encoder-decoder (ED)} \citep{attention_is_all, t5, bart} and \emph{Decoder-only (DO)} \citep{gpt2, opt}. 

ED models, which consist of two components (an encoder and a decoder), process inputs and targets (outputs) independently, with different parameter sets: The encoder processes the inputs with self-attention layers and passes its output to the decoder. Then, the decoder autoregressively generates the target token by token by applying autoregressive-attention and cross-attention (with the output of the encoder). On the other hand, DO models consist of autoregressive-attention layers that process inputs and targets together. Typically, the target sequence is concatenated to the input sequence (sometimes, with a separation token between them, such as ``TL;DR'' for summarization). 

Notice that in contrast to the DO model, the encoder component represents each token of the input sequence by sharing information from all the tokens in the input (via self-attention), while the DO model represents an input token by sharing information only from its preceding tokens (via autoregressive-attention). Another difference between the two architectures is that each layer of the decoder component of the ED model, applies cross-attention to the target tokens by conditioning on the last hidden states of the input tokens. This is in contrast to the decoder layers of the DO model which apply autoregressive-attention to the target inputs by conditioning on the same layer hidden states of the input tokens.

ED and DO models differ not only in the architecture but also in the pre-training objectives. Whereas DO models are trained with an autoregressive language modeling objective (given previous tokens, predict the following one), ED models are trained with a masked language modeling objective (given a sequence with masked spans, predict the missing tokens). 

As a result of these differences (encoder component, attention mechanisms, and training objectives), the models exhibit different inductive biases, which affect their performance. While ED models are more popular for classification, summarization, and NMT tasks, DO models excel on open-text generation and zero-shot or few-shot learning \citep{t5, zero_shot_arch}. Furthermore, the two architectures have different computational complexities (see the discussion in the next subsection, \S\ref{sub:complexity}). Nonetheless, the increasing popularity of huge DO models like GPT-3/4 and PaLM \citep{gpt3, palm, gpt-4}, which have impressive generation capabilities, has led to the question of \emph{``whether ED models are still relevant for NLG''}, a question that we aim to answer in the first stage of our study. 

To build an NLG system, it is necessary to select an architecture that meets its needs. In the spirit of our realistic setup, we compare various off-the-shelf ED and DO LMs from different families, and show that ED models outperform DO models in conditional generation tasks. These findings are in line with the recent work of \citet{ul2}, which in contrast to us, trained from scratch LMs. For the DO architecture, we use the GPT2-family models \citep{gpt2}: \gpts , \gptm , and \gptl ; and the recent OPT-family models \citep{opt}: \opts and \optm . For ED models we use the same models which are described in the main paper: T5-family models \citep{t5}: \tfives{} and \tfivel ; and the BART-family models \citep{bart}: \bartb{} (base version) and \bartl .


 \begin{figure}[t]
    \centering
    \includegraphics[width=0.48\textwidth]{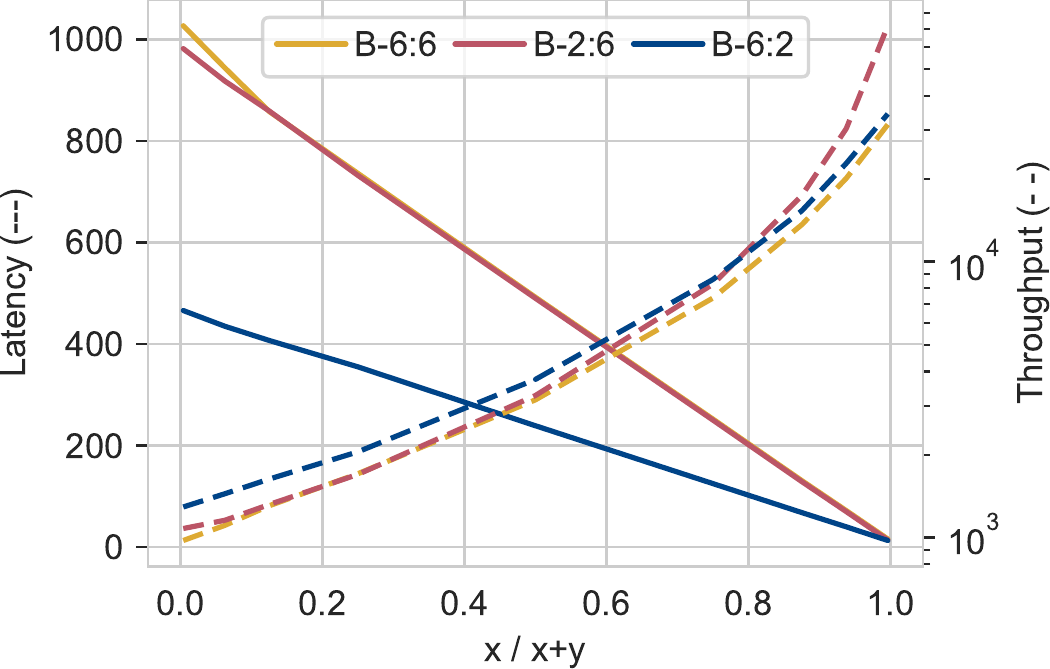}
    \caption{Comparing different pruning strategies for BART-E:D model, where E and D are the numbers of encoder and decoder layers respectively. X-axis is the ratio of the input length to the sum of the input and target lengths (=256). Left: Latency in milliseconds (solid line), Right: Throughput -- examples per minute in a log scale (dashed line). }
    \label{fig:pruning}
\end{figure}

\subsection{Complexity Analysis}
\label{sub:complexity}

For the theoretical complexity analysis, we assume all the transformer models have the same hidden size dimension and ignore it in our analysis. We consider two types of models: ED with $E$ encoder layers and $D$ decoder layers, and a DO model with $D$ decoder layers. The input and target lengths are $m$ and $n$, respectively. For decoding, we assume that hidden states of previously generated tokens are cached and re-used (i.e., for the $i$-th token, the decoder layers perform operations only for it). We do not discuss space complexity, as it depends on the exact implementation \citep{no_n2} and memory utilization of the device. Therefore, we do not connect the throughput measure to theoretical analysis.

A single encoder layer has a quadratic time complexity in terms of input length $O(m^2)$. An ED decoder layer consists of a causal-attention and cross-attention and therefore has a time complexity of $O(n(m+n)$. Thus, ED model has a complexity of $O(m^2E+n(m+n)D)$. Since we concatenate the input and the target for DO models, a single decoder layer of a DO has a time complexity of $O((m+n)^2)$. Thus, a DO model has a complexity of $O(m^2D+n(m+n)D)$. This suggests that an ED model can have the same time complexity as a DO model (when $E=D$) while having double parameters because the encoder and the decoder layers do not share parameters (excluding cross-attention weights, which account for a small portion of the total weights \citep{t5}).

Note that the number of floating-point operations (FLOPs, see subsection \S\ref{sub:evaluation}), is compatible with the theoretical complexity. As a result, it is possible to verify the observation above -- an ED model with double the number of layers and parameters as a DO model should have roughly the same number of FLOPs. Consider Table~\ref{sub:architecture} and take for example \gptm{}, which has 24 decoder layers, and compare it to \tfivel{} which has 48 layers (both of them have the same number of heads and the same hidden dimension, see Table~\ref{tab:arch}). Indeed, they have the same number of FLOPs. 
On the other hand, there are differences in the practical measures. While the latency of \gptm{} is smaller than \tfivel{}, its memory footprint is larger, which results in smaller throughputs. This highlights the complex nature of the connection between the theoretical and the practical measures (e.g., FLOPs and latency), which depend on the device, implementation, and hardware utilization that enable parallelism.

Now, compare models from the same family but with different sizes. As can be seen, the ratio between the latencies of the models does not reflect the large compression rate between their sizes. For example, \tfivel{} is ~12 times larger than \tfives{}, however, it is only 3.7 times faster. Likewise, \gptl{} is ~6 times larger than \gpts{}, but is only 2.9 times faster. On the other hand, the throughput reflects much better the size differences. 
This demonstrates the complex relationship between architectural decisions and computational measurements and suggests that architectural decisions should be taken according to the (specific) task and system needs. 

We next present a big $O$ notion when assuming that operations can be parallelized (as in GPUs). This notation reflect better the latency: a practical measure of the time for generating a single target example. With full parallelism, the complexity of processing the input in a single encoder layer (for ED) is reduced from $O(m^2)$ to $O(m)$ (see \citet{deep_shallow}). The same is true for the DO decoder layer when it processes the input since it is capable of processing all of it at the same time. However, since the target is generated by one token at a time (autoregressive), the processing complexity in each layer that processes the target remains $O(n(m+n))$. As a result, the time complexity of ED is $O(mE+n(m+n)D)$ and of DO is $O(mD+n(m+n)D)$, which is equal to the ED complexity when $E=D$. Nevertheless, there are differences in practical measurements. 

The theoretical analysis when allowing parallelism sheds light on two observations that come up from the practical analysis. The first one is that the length of the target has a higher impact on the latency, than the length of the input. This is expected in the autoregressive generation process, where the relationship between the complexity and the input length is linear, while quadratic for the target length. This is supported by Table~\ref{tab:compute}: for all models, altering only the input size minimally affects the latency. 

The second observation is pruning decoder layers has a higher impact on the latency than pruning encoder layers. This is also expected since each decoder layer contributes $O(n(m+n))$ to the total latency complexity, whereas a single encoder layer contributes $O(m)$. This is verified in Table~\ref{tab:compute} and in Figure~\ref{fig:pruning}: the encoder pruned model, \bartp{}, has roughly the same latency as its full version, \bartb{}. Conversely, the decoder pruned model \barts{} has a smaller latency from both. 

The behavior of throughput is more complex than latency. While the pruned decoder model consistently has a smaller latency regardless of input length (as shown in Figure~\ref{fig:pruning}), the pruned encoder (\bartp{}) has a higher throughput than the pruned decoder (\barts{}) for longer inputs, as indicated by the crossover at around 0.8 on the X-axis in Figure~\ref{fig:pruning}.

\begin{table}
\centering
\begin{adjustbox}{width=0.5\textwidth}
\begin{tabular}{llccccc}
\toprule
\textbf{Arch.} & \textbf{Model} & \textbf{Enc.} & \textbf{Dec.} & \textbf{Heads} & \textbf{Hidden} & \textbf{Params} \\
\midrule
DO & \gptl & 0 & 36 & 20 & 1280 & 774 \\
 DO & \gptm & 0 & 24 & 16 & 1024 & 354 \\
 DO & \gpts & 0 & 12 & 12 & 768 & 124 \\
 DO & \optm & 0 & 24 & 16 & 1024 & 331 \\
 DO & \opts & 0 & 12 & 12 & 768 & 125 \\
 ED & \tfivel & 24 & 24 & 16 & 1024 & 737 \\
 ED & \tfives & 6 & 6 & 8 & 512 & 60 \\
 ED & \bartl & 12 & 12 & 16 & 1024 & 406 \\
 ED & \bartb & 6 & 6 & 16 & 768 & 139 \\
 ED & \barts & 6 & 2 & 16 & 768 & 111 \\
 ED & \bartp & 2 & 6 & 16 & 768 & 101 \\
\bottomrule
\end{tabular}
\end{adjustbox}
\caption{Architectural details for different models. Enc. and Dec. present the number of layers. Heads present the number of attention heads at each layer. Hidden is the size of the hidden dimension and Params is the number of parameters in millions.}
\label{tab:arch}
\end{table}

\subsection{Task Performance Analysis}
\label{sub:performance}

In this subsection, we discuss the differences in task performance between off-the-shelf ED and DO models, which are finetuned on our four datasets. The average results (over the four tasks) are provided in Table~\ref{tab:performance}. For all datasets and models, ED models outperform DO models. Presumably, a better inductive bias is injected to the ED models: (1) By applying self-attention (and not autoregressive-attention) to the conditioned input sequence; (2) By the fact that in contrast to the DO model, the decoder component of the ED model attends to the last hidden states of the conditioned input sequence from its first layer. This is unlike the DO model, where each layer applies attention to hidden states of the same layer. 

Our results for conditional generation tasks in a finetuning setup are in line with other works \citep{t5, ul2} which trained LMs from scratch. This finding is particularly relevant for NLP practitioners who aim to develop a specialized in-house model for a specific NLG task. Our findings also raise the question of why huge language models, such as GPT-3 and PaLM \citep{gpt3, palm} are DO, and \citet{zero_shot_arch} answer it by showing that DO models excel in zero and few-shot setups. Indeed, in the final part of our study, which involves an extreme setup where labeled data is unavailable, we use GPT-4, a Decoder-only model with zero-shot capabilities, to generate PTs. 
Our equivocal results lead us to continue only with ED models (T5 and BART) for our compression study (stages 1-8).

\section{KD without Labeled Data}
\label{sec:extreme}

In the final phase of our study, we intend to explore the possibility of scaling up our experimental setup. This is accomplished by working with only a limited number of unlabeled examples and without any labeled examples. We refer to this setup as \emph{extreme setup}.
It is important to note that unlike the realistic setup, which incorporates a medium-sized labeled dataset, the extreme setup poses a challenge for fine-tuning a teacher model due to the lack of labeled examples. In that case, we need to utilize as our teacher a huge LM, such as GPT-4, which has zero-shot and few-shot capabilities and can generate plausible PTs. 

The main goal of this phase is to investigate the transferability of the KD conclusions from our realistic setup to the extreme setup, which possess the following differences since it involves a huge zero-shot LM as the teacher: (1) The teacher is a Decoder-only model (since this is the standard architecture for zero-shot and few-shot LMs) and the student is an Encoder-decoder model (following our findings that they outperform Decoder-only models, see \S\ref{sec:results}); (2) The teacher and the student have different tokenizers, which means that a sequence-alignment algorithm is needed to perform Logits KD; (3) Unlike the realistic setup where the computational training cost could be neglected, in the extreme setup we assume that querying the huge teacher is financially costly and therefore we limit its usage. 

The third difference above impacts the design choice of the extreme setup, and we limit the number of unlabeled data to a few thousand. 
In addition, we do not consider the Joint-Teaching method due to its high cost compared to other methods. This is because it requires querying the teacher every time we generate a PT with the student (to extract the teacher logits).
However, notice that extracting the logits of GPT-4 and generating multiple PTs is approximately equivalent to generating a single PT. This is because the prompt and the input, consisting of many tokens, are processed only once, and the marginal cost of generating multiple (relatively short) PTs is low.

\subsection{Experimental Setup}
\label{sub:extreme_setup}

\smallskip\noindent\textbf{Models and datasets} We utilize GPT-4 as our teacher model and \tfives{} as the student model. For generating PTs with GPT-4, we use a prompt that contains a task instruction and three demonstrations of labeled examples (few-shot learning). 

We consider two NLG tasks: (1) Question Generation -- we use the \squad{} dataset and sample 3000, 250 and 500 examples as the train, development and test, respectively; (2) Simplification and style transfer -- we use the \shakespeare{} dataset and sample 1500, 250 and 350 examples as the train, development, and test, respectively. Notice that both the training and development sets do not contain any labeled data. Only the test set includes labeled data, which is used for evaluation purposes.

\smallskip\noindent\textbf{Methods and baselines} We present the test results for the following baselines and methods: (a) The GPT-4 teacher; (2) A \tfives{} model which is trained using ground-truth (GT) targets to compare with the GPT-4 teacher; (c) Student fine-tuning with a single PT; (d) Student fine-tuning with multiple PTs; (e) Student training with a single PT and Logits KD; (f) Student training with multiple PTs and Logits KD.

We train each model (except GPT-4) using four learning rates: [0.003, 0.001, 0.0005, 0.0003]. In addition, as we explain \S\ref{sub:extreme_results}, we also include results when we train the models of (c)-(f) using golden targets (ground-truth labels) for the development set.

\begin{table*}
\centering
\begin{adjustbox}{width=0.98\textwidth}
\begin{tabular}{l|cccc|cccc|cccc|cccc}
\toprule
  & \multicolumn{8}{c|}{dev contains GTs} & \multicolumn{8}{c}{dev contains PTs} \\

  & \multicolumn{4}{c|}{\squad{}} & \multicolumn{4}{c|}{\shakespeare{}}  & \multicolumn{4}{c|}{\squad{}} & \multicolumn{4}{c}{\shakespeare{}} \\
  \midrule
\textbf{Method} & \textbf{BL}  & \textbf{RG}  & \textbf{BS}  & \textbf{PP} & \textbf{BL}  & \textbf{RG}  & \textbf{BS}  & \textbf{PP} & \textbf{BL}  & \textbf{RG}  & \textbf{BS}  & \textbf{PP} & \textbf{BL}  & \textbf{RG}  & \textbf{BS}  & \textbf{PP}\\
\midrule
 a. GPT-4 (Teacher) & 13.6 & 37.6 & 75.0 & & 21.4 & 42.3 & 79.4 &  &  &  &  &  &  &   &  &   \\
 b. \tfives{} + GT labels & 17.8 & 36.4 & 75.2 & 2.1 & 21.2 & 42.1 & 76.4 & 2.1 & & & & &  &  &   \\
 \midrule
 c. \tfives{} + PT & 11.3 &       31.9 &              72.0 &    2.49 &           \textbf{19.1} &       \textbf{41.1} &              76.0 &    2.37 &           11.7 &       32.6 &              72.4 &    2.65 &           18.9 &       \textbf{41.1} &              76.0 &    2.47 \\
 d. \tfives{} + PTs & 11.4 &       32.2 &              72.0 &     2.4 &           19.0 &       \textbf{41.1} &              \textbf{76.1} &    \textbf{2.36} &           11.6 &       32.7 &              72.4 &    2.84 &           \textbf{19.2} &       \textbf{41.1} &              \textbf{76.1} &    \textbf{2.36} \\
 e. \tfives{} + Logits + PT &11.3 &       32.0 &              72.1 &    2.51 &           18.4 &       40.7 &              75.8 &    2.47 &           11.6 &       32.4 &              72.4 &    2.72 &           18.5 &       41.0 &              75.9 &    2.66 \\
 f. \tfives{} + Logits + PTs & \textbf{12.0} &       \textbf{32.7} &              \textbf{72.5} &     \textbf{2.4} &           18.9 &       41.0 &              75.9 &    2.42 &           \textbf{11.9} &       \textbf{32.8} &              \textbf{72.5} &    \textbf{2.48} &           19.1 &       40.7 &              75.0 &    2.52 \\

\bottomrule
\end{tabular}
\end{adjustbox}
\caption{BLEU (BL), ROUGE (RG), BERTScore (BS), and PPL (PP) test scores of different baselines and KD methods in the extreme setup. The four left columns present the scores when the development set contains ground-truth (GT) targets, and the four right when it contains PTs generated by the GPT-4 teacher.}
\label{tab:kd_gpt_full}
\vspace{-1em}
\end{table*}
\begin{figure}[t]
    \centering
    \includegraphics[width=0.5\textwidth]{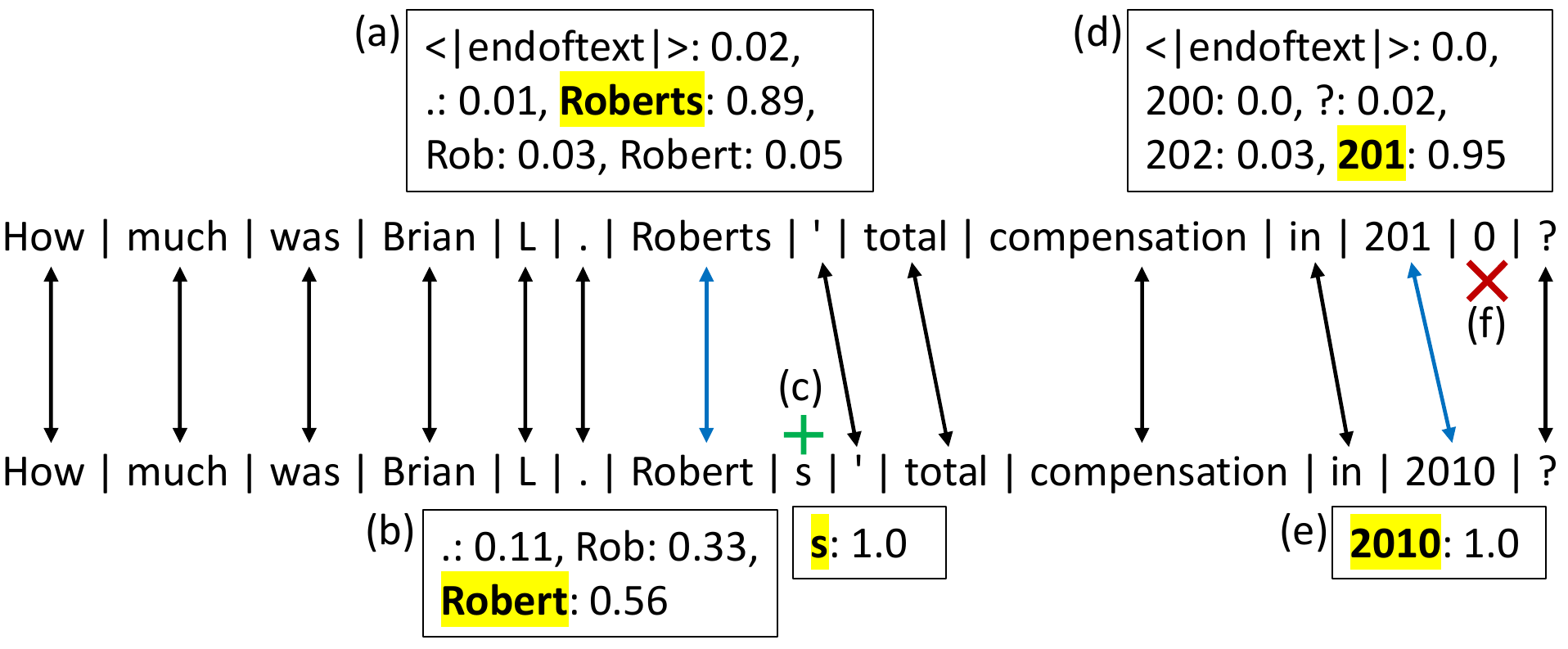}
    \caption{An example of an alignment between the teacher's tokens (top) and the student's tokens (bottom). Black double arrows represent an exact match while blue arrows represent replacement. Green $+$ represents an insertion while red $\times$ represents a deleted token. The boxes present the probabilities of the next token in the location of the yellow highlighted token.}
    \label{fig:alignment}
    \vspace{-1em}
\end{figure}

\smallskip\noindent\textbf{Tokenizers Alignment} In the extreme setup, the teacher is a Decoder-only model (GPT-4), and the student is an Encoder-decoder model (\tfives) that does not share the same tokenizer. Therefore, to perform Logits KD, where the probabilities of the next token's logits are used for distillation, two types of token alignment are required: (1) Matching each token in the teacher's tokenized PT sequence with its corresponding token in the student's tokenized PT sequence; (2) Matching the tokens from the teacher's logits to tokens from the student's vocabulary.

For example, consider the black and blue arrows in Figure~\ref{fig:alignment}. These arrows demonstrate the first type of match, where we align the tokens of the tokenized PT sequences. Additionally, some tokens might be inserted (c) or deleted (f). Similar to \citet{specializing}, we use the well-known dynamic programming Needleman–Wunsch algorithm \citep{needleman} for sequence alignment to find this mapping. The output of the algorithm is a sequence of edit operations: match, replacement, insertion, and deletion. We consider two tokens as a match if the algorithm determines them as such or if they are replaced and one is a prefix of the other. For instance, the blue arrows in Figure~\ref{fig:alignment} represent a match via replacement.

The OpenAI API allows us to extract only the probability distribution over the top five tokens at each decoding step. However, their probability is usually close to 1. We align the top five tokens to the student's vocabulary by performing an exact match. Then, we apply softmax to the logits to make their probabilities sum to one. For example, (a) and (b) in Figure~\ref{fig:alignment} present such an alignment. Notice that some of the tokens of the teachers are omitted (``Robsert'' and ``|<endoftext|>''). In case the student token does not have a match in the teacher's top five tokens, we determine its probability as one. For example, (c) and (e) in Figure~\ref{fig:alignment}: (c) ``s'' is a token that is inserted and therefore its probability is one; and (e) the token ``2010'' does not appear in the top five tokens (d), and therefore its probability is one.  

The second type of alignment we need to perform is matching the teacher's logits of the next token prediction to tokens from the student's vocabulary. The OpenAI API allows us to extract only the probability distribution over the top five tokens at each decoding step. However, the sum of their probabilities is usually close to 1. We align the top five tokens with the student's vocabulary by performing an exact match. Then, we apply softmax to the logits to ensure their probabilities sum up to one. For example, (a) and (b) in Figure~\ref{fig:alignment} demonstrate such an alignment. Note that some tokens from the teacher are omitted (e.g., "Roberts" and "|<endoftext|>"). 
If the student's token does not have a match in the teacher's top five tokens, we assign its probability as one. For instance, in Figure~\ref{fig:alignment}, (c) "s" is an inserted token, so its probability is one; and (e) the token "2010" does not appear in the top five tokens (d), hence its probability is one.

\subsection{Results}
\label{sub:extreme_results}

In Table~\ref{tab:kd_gpt_full}, we present the results of the extreme setup. We do not include computational performance metrics as OpenAI does not detail the exact architecture of GPT-4. 
We find the results vary greatly between different initializations and learning rates. The observed difference in performance can be primarily attributed to the unique extreme setup. The limited number of training instances and the distinct distribution of PTs, generated by GPT-4 rather than a fine-tuned model contribute to this variation. Additionally, the discrepancy between the development set, consisting of PTs, and the test set, containing ground-truth targets, negatively affect model selection \citep{specializing}. 

To address the issue of variability, we present the average scores over different learning rates. Additionally, we include the results from experiments conducted with development sets that contain ground-truth (GT) targets (as shown in the four left columns of Table~\ref{tab:kd_gpt_full}).
Indeed, the correlation between the development score and the test score is considerably low (0.06 and 0.12 for the \squad{} and \shakespeare{} datasets, respectively) when the development sets are PTs, in contrast to the higher correlations observed when the development sets are GTs (0.57 and 0.66). 

As depicted in Table~\ref{tab:kd_gpt_full}, the overall trends in the left four columns (development set with GTs) align with those in the right four (development set with PTs). 
The results are mixed: for the \squad{} dataset,  Logits KD with multiple PTs outperforms the other methods, which is in line with the conclusions from the realistic setup.
Surprisingly, incorporating Logits KD has a positive effect only when there are multiple PTs. 
In the \shakespeare{} dataset, Logits KD does not improve the student. We believe this is due to the difficulties with aligning the tokenizers and call for further research.
Nevertheless, another conclusion from the realistic setup, which also holds in the extreme setup, is that generating multiple PTs is preferable over a single PT. 
\section{Additional Implementation Details}
\label{sec:additional_implementation}
Our experiments are conducted in the PyTorch framework. Models are trained on a machine equipped with 4 Nvidia Tesla v100 GPUs (for \xsum{} and \squad{}; in that case, we use DDP training) or with Nvidia GeForce RTX 4080 (for \art{} and \shakespeare{}). 

\paragraph{Training} We optimize our model with the AdamW optimizer, with a weight decay of $1e-5$, $\varepsilon=1e-8$, $100$ warmup steps, and a linear learning rate scheduler. We use the largest batch size that fits the GPU for every dataset and model. However, for a fair comparison, we accumulate the gradients and update the model every 96 training examples for any experiment (same number of gradient updates). For BART models we apply half-precision training and inference. 

The validation metric for \xsum{} is ROUGE-2 (F1) and for \squad{}, \art{} and \shakespeare{} is \bleu{}. In addition, for \xsum{} we use ``summarize:'' as a prefix for T5 models and ``TL;DR'' as a suffix for DO models. For \squad{} we use ``ask:'' as a prefix and suffix for T5 and DO models respectively. For \art{} we use ``explain:'' as a prefix and suffix for T5 and DO models respectively, and for \shakespeare{}, we use ``modern:'' as a prefix and suffix for T5 and DO models respectively.

\paragraph{Fine-tuning} We examine multiple learning rates for each model and dataset as follows: for our student models, \tfives{} and \bartb{}, and smaller Decoder-only models, \gpts , \opts --we search within 8 different learnings rates in the range of $[5e-2, 1e-5]$ and train the models for 35 epochs. For our teacher models, \tfivel{} and \bartl{} and the remaining decoder-only models--6 learning rates in $[5e-3, 1e-6]$ and 20 epochs. For fine-tuning the decoder-only models, we concatenate the input and the target, separated by a task suffix, and calculate the loss only on the target tokens.

\paragraph{Evaluation} We evaluate every model two times at each epoch (at the middle and the end) and select the best checkpoint according to the development set performance (see \S\ref{sub:datasets} for more details about the measure used in each dataset). For computational reasons, when we evaluate the model on the development set, we generate predictions for no more than 1K. We use DeBERTa-base model \citep{deberta}, fine-tuned on the MNLI dataset as the backbone model for calculating \bertscore s.

\paragraph{Knowledge Distillation} For computational reasons, the learning rate which is used for training the student is selected according to the development performances in the fine-tuning stage (and reported on Tables~\ref{tab:XSUM}, \ref{tab:SQuAD}, \ref{tab:ART}, \ref{tab:Shakespeare}). Since we have observed that the convergence time of the students in KD setups is slow (unlike fine-tuning), we train the models for 192 epochs but stop the training if there is no improvement in the performance for 16 epochs (32 evaluation steps). We note that all of the experiments were stopped before the last epoch. Following \citet{xtremedistil, autodistil}, we perform a fine-tuning stage for 10 epochs after selecting the best checkpoint during the KD stage. The final checkpoint is selected either from the KD or fine-tuning stages.

For Logits KD we minimize the KL divergence between the student and the teacher logits. We also tried using Label-Smoothing \citep{label_smoothing} with different scaling temperatures, however, it did not help the distillation. For Sequence-Level KD we fine-tune the student on pseudo-targets generated with beam search in addition to the original ground truth targets. For Noisy KD we only apply noise to the teacher's logits (as it is shown to be more important than applying noise to the input, and since we don't focus on input manipulations in this study). For Attention-Relations KD we distill relations from the last encoder and last decoder layers and scale the weights of the loss components to 1 at the start of the training. 

\paragraph{Pseudo Targets} For generating pseudo targets we use nucleus sampling with $P=0.95$, for high temperature sampling we use $\tau=1.5$.
When generating pseudo targets with the teacher using beam search, we use a beam size of 16. For sampling, we generate 48 pseudo targets (these are the largest sizes that fit on v100 GPU for \xsum{}). 

In experiments with a single pseudo target, we select the highest-ranked prediction among the generated targets using beam search with a beam of size 16.
We augment the training data with PTs by adding pairs of input and \textbf{a single} PT for each labeled or unlabeled example (depending on the experiment). In experiments with multiple PTs, we use a different single PT at every epoch (alternatively, the student could learn from multiple pseudo targets of the same input on every epoch). We do it for two main reasons: first, we want a fair comparison between experiments with single or multiple pseudo targets. Second, we have observed that the ground truth of the labeled data is important--this way, the student sees more of it as the proportion of ground truth targets is larger than the alternative. 

We use nucleus sampling and for high temperature sampling we use $\tau=1.5$. For computational reasons, we generate all the teacher PTs once and reuse them. Conversely, the student PTs are generated on-the-fly, since the student is continuously updated during training. In Joint-Teaching, we generate PTs with the student in 50\% of the training steps (in the remaining 50\% we use the teacher). 

\paragraph{Computational Profiling} All computational profiling experiments are conducted on Nvidia GeForce RTX 4080. Following \citet{deci}, we do a GPU warmup for 10 steps and then average 100 computational measurements. For every dataset, we use the maximum input and target length as reported in Table~\ref{tab:datasets}. FLOPs are measured for a full forward step. Latency and memory footprint are measured for generating a single example. For measuring Throughput, which is the maximum number of examples the model can process in a minute, we find the maximum batch size that does not exceed 16GB during the generation, and then measure the throughput.

\subsection{The \shakespeare{} Dataset}
\label{sub:shakespeare}
We construct a new dataset for the well-explored style transfer task (which is also a simplification task) of translating Shakespeare's texts to modern English. We combined three existing datasets: two parallel datasets of Shakespeare’s original texts and their modern versions \citep{shakespeare, shakespeare_st} and a third dataset containing only unlabeled texts from Shakespeare’s plots that are not part of the other two datasets \citep{shakespeare_unsupervised}.
A particular advantage of this dataset is that it consists of publically available datasets, while many other datasets for thesimplification task are not public. Moreover, in this dataset we have access to both labeled (original alongside modern texts) and unlabeled (original texts) data. Additionally, the labels (modern English texts) are of very high quality as experts produce them. Finally, the task of this dataset is harder than other style transfer and simplification cases since the difference between the original text and the simplified version is not limited to a small number of words. We hope this dataset will contribute to the NLP community.

\subsection{URLs of Code and Data}
\label{sub:urls}

\begin{itemize}
  \item \textbf{Code Repository} - code and datasets: \texttt{\href{https://github.com/nitaytech/KD4Gen}{github.com/nitaytech/KD4Gen}}.
  \item \textbf{HuggingFace} \citep{huggingface} - code and pretrained weights for language models, tokenizers, and datasets: 
  \texttt{\href{https://huggingface.co/}{huggingface.co/}}.
  \texttt{\href{https://huggingface.co/docs/accelerate/index}{huggingface.co/docs/accelerate/index}}.
  \item \textbf{Torchprofile} - for measuring FLOPs: 
  \texttt{\href{https://github.com/zhijian-liu/torchprofile}{github.com/zhijian-liu/torchprofile}}.
\end{itemize}

\section{Additional Results}
\label{sec:additional_results}

In this section, we report additional details and the complete results of our study. Table~\ref{tab:arch} presents a full description of the architecture of the models used in our study. Table~\ref{tab:compute} provide full results of our computational measurements. 
In Tables~\ref{tab:XSUM}, \ref{tab:SQuAD}, \ref{tab:ART} and \ref{tab:Shakespeare} we report on the performances of every experiment we conduct, for \xsum{}, \squad{}, \art{} and \shakespeare{} datasets, respectively. 

\begin{table}[h!]
\centering
\begin{adjustbox}{width=0.5\textwidth}
\begin{tabular}{c|lccccc}
\toprule
 \textbf{Dataset} & \textbf{Model} & \textbf{FLOPs} & \textbf{Lat.} & \textbf{G. Mem} & \textbf{TP} & \textbf{Batch} \\
\midrule
 \multirowcell{11}{XSUM\\40K\\$|x|=480$\\$|y|=32$} & \gptl & 84.0 & 630 & 539 & 555 & 23 \\
 & \gptm & 38.8 & 424 & 345 & 1177 & 42 \\
 & \gpts & 13.6 & 211 & 198 & 3335 & 78 \\
 & \optm & 36.3 & 344 & 294 & 1415 & 49 \\
 & \opts & 13.6 & 170 & 180 & 3728 & 86 \\
 & \tfivel & 38.7 & 539 & 112 & 1297 & 116 \\
 & \tfives & 2.7 & 144 & 30 & 13392 & 525 \\
 & \bartl & 19.6 & 254 & 59 & 3349 & 243 \\
 & \bartb & 5.8 & 132 & 36 & 8361 & 428 \\
 & \bartp & 2.8 & 131 & 36 & 12320 & 432 \\
 & \barts & 5.1 & 68 & 36 & 10025 & 433 \\\cmidrule{1-7}
 \multirowcell{11}{SQuAD\\17.5K\\$|x|=320$\\$|y|=32$} & \gptl & 56.7 & 604 & 361 & 849 & 35 \\
 & \gptm & 26.1 & 410 & 223 & 1804 & 65 \\
 & \gpts & 9.2 & 215 & 124 & 5128 & 124 \\
 & \optm & 24.4 & 333 & 193 & 2143 & 76 \\
 & \opts & 9.2 & 167 & 112 & 5748 & 138 \\
 & \tfivel & 26.1 & 530 & 77 & 2011 & 166 \\
 & \tfives & 1.8 & 143 & 16 & 22256 & 984 \\
 & \bartl & 13.3 & 250 & 41 & 4759 & 350 \\
 & \bartb & 3.9 & 133 & 16 & 11066 & 956 \\
 & \bartp & 2.0 & 131 & 16 & 15000 & 963 \\
 & \barts & 3.4 & 67 & 16 & 13009 & 965 \\\cmidrule{1-7}
 \multirowcell{11}{ART\\10K\\$|x|=48$\\$|y|=32$} & \gptl & 12.5 & 582 & 58 & 4417 & 219 \\
 & \gptm & 5.7 & 401 & 34 & 9421 & 428 \\
 & \gpts & 2.0 & 208 & 18 & 26177 & 824 \\
 & \optm & 5.3 & 325 & 32 & 10328 & 458 \\
 & \opts & 2.0 & 161 & 17 & 30164 & 911 \\
 & \tfivel & 5.9 & 533 & 22 & 10661 & 581 \\
 & \tfives & 0.5 & 142 & 3 & 109777 & 5101 \\
 & \bartl & 3.2 & 250 & 12 & 13704 & 1197 \\
 & \bartb & 1.0 & 132 & 5 & 21259 & 3088 \\
 & \bartp & 0.8 & 131 & 5 & 22818 & 3111 \\
 & \barts & 0.8 & 67 & 2 & 23408 & 7359 \\\cmidrule{1-7}
 \multirowcell{11}{Shakespeare\\7K\\$|x|=48$\\$|y|=48$} & \gptl & 15.0 & 883 & 70 & 3129 & 182 \\
 & \gptm & 6.9 & 600 & 38 & 6849 & 383 \\
 & \gpts & 2.4 & 306 & 18 & 19237 & 824 \\
 & \optm & 6.4 & 484 & 38 & 6439 & 386 \\
 & \opts & 2.4 & 241 & 17 & 21902 & 911 \\
 & \tfivel & 7.2 & 789 & 28 & 7422 & 453 \\
 & \tfives & 0.6 & 212 & 3 & 75309 & 4063 \\
 & \bartl & 3.9 & 367 & 15 & 9200 & 958 \\
 & \bartb & 1.3 & 192 & 6 & 13251 & 2562 \\
 & \bartp & 1.1 & 191 & 6 & 13859 & 2581 \\
 & \barts & 1.0 & 96 & 3 & 14836 & 5197 \\
\bottomrule
\end{tabular}
\end{adjustbox}
\caption{Computational profiling results. For each dataset, we use the maximum input and target lengths as appear in table \ref{tab:datasets}. FLOPs (in billions) are the number of floating-point operations required for a full forward step. Latency (Lat.) (in milliseconds) is the time required for generating a single example. G. Mem is the memory footprint in MB for generating a single example. Throughput (TP) is the number of examples that can be generated per minute when taking the maximum batch size (Batch) that does not exceed 16GB.}
\label{tab:compute}
\end{table}

\begin{table*}
\centering
\begin{adjustbox}{width=\textwidth}
\begin{tabular}{llllllllllllllllll}
\toprule
 \textbf{Model} & \textbf{Obj.} & \textbf{PT In.} & \textbf{Decoding} & \textbf{PT St.} & \textbf{LR} & \textbf{FT} & \textbf{Dev} & \textbf{BLEU} & \textbf{ROUGE} & \textbf{PPL} & \textbf{R1} & \textbf{R2} & \textbf{RL} & \textbf{BS-F1} & \textbf{BS-P} & \textbf{BS-R} & \textbf{MET} \\
\midrule
 \tfivel & FT & - & - & - & 5e-5 & F & 17.7 & 11.5 & 29.3 & 1.7 & 39.0 & 16.9 & 31.9 & 72.7 & 73.8 & 71.8 & 33.5 \\
 \tfives & FT & - & - & - & 3e-3 & T & 12.2 & 7.6 & 23.2 & 3.1 & 32.4 & 11.5 & 25.9 & 68.3 & 69.0 & 67.9 & 26.8 \\
 \tfives & FT & L & 1-BS & T & 3e-3 & F & 13.5 & 8.5 & 24.6 & 2.9 & 33.7 & 12.8 & 27.3 & 69.2 & 70.4 & 68.3 & 27.9 \\
 \tfives & Noisy & - & - & - & 3e-3 & F & 13.7 & 8.3 & 24.9 & 2.2 & 34.2 & 12.9 & 27.6 & 69.7 & 71.3 & 68.4 & 28.0 \\
 \tfives & Att-Rel & - & - & - & 3e-3 & F & 14.0 & 8.8 & 25.4 & 2.3 & 34.8 & 13.4 & 28.1 & 70.2 & 71.5 & 69.1 & 28.8 \\
 \tfives & Logits & - & - & - & 3e-3 & F & 14.1 & 8.5 & 25.1 & 2.3 & 34.4 & 13.1 & 27.8 & 69.9 & 71.4 & 68.7 & 28.3 \\
 \tfives & Logits & L & 1-BS & T & 3e-3 & F & 14.2 & 8.4 & 25.0 & 2.3 & 34.3 & 13.1 & 27.8 & 69.8 & 71.2 & 68.6 & 28.1 \\
 \tfives & Logits & L+U & 1-BS & T & 3e-3 & F & 15.8 & 9.9 & 27.1 & 2.1 & 36.4 & 15.0 & 29.8 & 71.0 & 69.9 & 72.4 & 30.5 \\
 \tfives & Logits & L+U & K-BS & T & 3e-3 & F & 15.9 & 10.3 & 27.3 & 2.1 & 36.8 & 15.3 & 29.9 & 71.2 & 70.3 & 72.2 & 31.1 \\
 \tfives & Logits & L+U & Samp. & T & 3e-3 & F & 16.3 & 10.5 & 27.9 & 1.9 & 37.3 & 15.7 & 30.6 & 71.8 & 70.5 & 73.3 & 31.5 \\
 \tfives & Logits & L+U & H-Samp. & T & 3e-3 & T & 16.3 & 10.5 & 27.9 & 1.9 & 37.4 & 15.6 & 30.5 & 71.7 & 70.7 & 72.8 & 31.6 \\
 \tfives & Logits & L+U & Samp. & S & 3e-3 & T & 16.5 & 10.2 & 27.8 & 1.9 & 37.4 & 15.6 & 30.5 & 71.8 & 73.2 & 70.6 & 31.4 \\
 \tfives & Logits & L+U & Samp. & T+S & 3e-3 & F & 16.6 & 10.7 & 28.2 & 1.9 & 37.8 & 16.0 & 30.9 & 71.8 & 70.9 & 73.0 & 32.0 \\
 \bartl & FT & - & - & - & 1e-5 & F & 19.0 & 13.0 & 31.1 & 1.7 & 41.0 & 18.8 & 33.6 & 73.9 & 73.0 & 75.0 & 35.8 \\
 \bartb & FT & - & - & - & 5e-5 & F & 15.7 & 10.0 & 27.6 & 2.0 & 37.0 & 15.5 & 30.3 & 72.1 & 73.8 & 70.6 & 31.1 \\
 \bartp & FT & - & - & - & 5e-5 & F & 12.3 & 8.0 & 23.8 & 2.4 & 32.8 & 12.2 & 26.3 & 69.4 & 70.2 & 68.7 & 27.5 \\
 \barts & FT & - & - & - & 5e-5 & F & 15.4 & 9.3 & 26.5 & 2.7 & 35.8 & 14.6 & 29.2 & 71.0 & 72.6 & 69.6 & 29.9 \\
 \barts & FT & L & 1-BS & T & 5e-5 & F & 15.3 & 9.8 & 27.0 & 2.6 & 36.2 & 15.1 & 29.7 & 71.1 & 72.7 & 69.8 & 30.3 \\
 \barts & Noisy & - & - & - & 5e-5 & F & 16.1 & 9.8 & 27.6 & 2.2 & 37.0 & 15.6 & 30.3 & 71.7 & 73.6 & 70.0 & 30.8 \\
 \barts & Att-Rel & - & - & - & 5e-5 & F & 16.4 & 9.9 & 27.6 & 2.4 & 37.0 & 15.6 & 30.3 & 71.7 & 73.4 & 70.2 & 31.0 \\
 \barts & Logits & - & - & - & 5e-5 & F & 16.2 & 10.0 & 27.6 & 2.4 & 37.0 & 15.6 & 30.2 & 71.8 & 73.4 & 70.4 & 31.1 \\
 \barts & Logits & L & 1-BS & T & 5e-5 & F & 16.6 & 10.4 & 27.9 & 2.3 & 37.2 & 16.0 & 30.5 & 72.0 & 73.7 & 70.4 & 31.4 \\
 \barts & Logits & L+U & 1-BS & T & 5e-5 & F & 17.5 & 11.1 & 28.8 & 2.2 & 38.2 & 16.8 & 31.4 & 72.5 & 71.1 & 74.2 & 32.4 \\
 \barts & Logits & L+U & K-BS & T & 5e-5 & F & 17.7 & 11.5 & 29.4 & 2.2 & 38.8 & 17.3 & 32.0 & 72.8 & 71.4 & 74.5 & 33.2 \\
 \barts & Logits & L+U & Samp. & T & 5e-5 & F & 18.3 & 11.6 & 29.6 & 2.0 & 39.1 & 17.6 & 32.3 & 73.1 & 71.7 & 74.8 & 33.4 \\
 \barts & Logits & L+U & H-Samp. & T & 5e-5 & F & 17.9 & 11.3 & 29.4 & 2.0 & 38.8 & 17.3 & 32.0 & 73.0 & 71.4 & 74.8 & 32.9 \\
 \barts & Logits & L+U & Samp. & S & 5e-5 & F & 18.0 & 11.3 & 29.4 & 2.0 & 39.0 & 17.2 & 32.0 & 73.0 & 74.5 & 71.7 & 33.2 \\
 \barts & Logits & L+U & Samp. & T+S & 5e-5 & T & 18.3 & 12.3 & 30.2 & 1.9 & 39.9 & 18.0 & 32.8 & 73.5 & 74.6 & 72.5 & 34.7 \\
 \gptl & FT & - & - & - & 5e-6 & F & 11.9 & 7.8 & 22.5 & 1.9 & 30.7 & 11.9 & 24.9 & 65.6 & 67.2 & 64.2 & 25.6 \\
 \gptm & FT & - & - & - & 1e-4 & F & 11.2 & 6.8 & 20.6 & 2.1 & 28.7 & 10.4 & 22.9 & 63.9 & 65.1 & 62.9 & 23.7 \\
 \gpts & FT & - & - & - & 1e-3 & F & 8.0 & 4.6 & 17.6 & 2.6 & 25.0 & 7.8 & 19.9 & 62.0 & 63.8 & 60.5 & 19.9 \\
 \optm & FT & - & - & - & 1e-4 & F & 10.6 & 6.7 & 20.8 & 2.8 & 29.1 & 10.5 & 22.8 & 64.5 & 65.8 & 63.8 & 24.3 \\
 \opts & FT & - & - & - & 7e-5 & F & 11.5 & 7.0 & 22.5 & 2.2 & 31.1 & 11.4 & 24.9 & 67.9 & 69.5 & 66.5 & 25.2 \\
\bottomrule
\end{tabular}
\end{adjustbox}
\caption{Results for summarization task, \xsum{} dataset.}
\label{tab:XSUM}
\end{table*}
\begin{table*}
\centering
\begin{adjustbox}{width=\textwidth}
\begin{tabular}{llllllllllllllllll}
\toprule
 \textbf{Model} & \textbf{Obj.} & \textbf{PT In.} & \textbf{Decoding} & \textbf{PT St.} & \textbf{LR} & \textbf{FT} & \textbf{Dev} & \textbf{BLEU} & \textbf{ROUGE} & \textbf{PPL} & \textbf{R1} & \textbf{R2} & \textbf{RL} & \textbf{BS-F1} & \textbf{BS-P} & \textbf{BS-R} & \textbf{MET} \\
\midrule
 \tfivel & FT & - & - & - & 5e-5 & F & 22.0 & 22.2 & 42.3 & 1.3 & 50.6 & 29.3 & 46.8 & 77.9 & 77.8 & 77.7 & 48.9 \\
 \tfives & FT & - & - & - & 5e-4 & F & 19.4 & 19.1 & 38.3 & 1.9 & 46.6 & 25.3 & 43.2 & 76.1 & 75.8 & 75.8 & 44.6 \\
 \tfives & FT & L & 1-BS & T & 5e-4 & F & 19.7 & 18.8 & 38.7 & 2.8 & 46.8 & 26.0 & 43.3 & 75.7 & 75.5 & 76.1 & 45.5 \\
 \tfives & Noisy & - & - & - & 5e-4 & F & 20.3 & 20.2 & 39.6 & 1.7 & 47.8 & 26.6 & 44.4 & 76.4 & 76.7 & 76.4 & 45.9 \\
 \tfives & Att-Rel & - & - & - & 5e-4 & F & 20.4 & 20.1 & 39.5 & 1.7 & 47.7 & 26.5 & 44.3 & 76.4 & 76.6 & 76.4 & 45.8 \\
 \tfives & Logits & - & - & - & 5e-4 & F & 20.2 & 20.4 & 39.8 & 1.7 & 47.9 & 26.9 & 44.6 & 76.8 & 76.5 & 76.5 & 46.0 \\
 \tfives & Logits & L & 1-BS & T & 5e-4 & F & 19.9 & 19.6 & 39.4 & 1.8 & 47.7 & 26.5 & 44.0 & 76.2 & 76.3 & 76.5 & 46.1 \\
 \tfives & Logits & L+U & 1-BS & T & 5e-4 & F & 20.6 & 20.2 & 40.1 & 1.7 & 48.2 & 27.3 & 44.8 & 76.5 & 76.7 & 76.5 & 46.7 \\
 \tfives & Logits & L+U & K-BS & T & 5e-4 & F & 20.8 & 21.0 & 40.8 & 1.6 & 49.0 & 27.8 & 45.5 & 77.1 & 76.9 & 76.9 & 47.1 \\
 \tfives & Logits & L+U & Samp. & T & 5e-4 & F & 21.1 & 20.9 & 40.5 & 1.6 & 48.6 & 27.7 & 45.3 & 76.9 & 76.9 & 76.8 & 47.0 \\
 \tfives & Logits & L+U & H-Samp. & T & 5e-4 & F & 21.6 & 21.3 & 40.9 & 1.5 & 49.0 & 28.0 & 45.6 & 77.2 & 77.0 & 77.0 & 47.2 \\
 \tfives & Logits & L+U & Samp. & S & 5e-4 & F & 20.8 & 20.9 & 40.7 & 1.6 & 48.9 & 27.8 & 45.4 & 76.9 & 77.0 & 77.0 & 47.3 \\
 \tfives & Logits & L+U & Samp. & T+S & 5e-4 & F & 21.5 & 20.9 & 40.6 & 1.5 & 48.9 & 27.7 & 45.2 & 77.0 & 76.9 & 76.8 & 47.1 \\
 \bartl & FT & - & - & - & 1e-5 & F & 21.1 & 21.5 & 41.9 & 1.4 & 50.2 & 28.9 & 46.7 & 77.8 & 78.3 & 77.5 & 48.0 \\
 \bartb & FT & - & - & - & 1e-4 & F & 18.4 & 19.3 & 39.2 & 1.7 & 47.7 & 26.1 & 43.8 & 76.3 & 76.3 & 76.5 & 45.9 \\
 \bartp & FT & - & - & - & 1e-4 & F & 12.3 & 11.8 & 28.7 & 1.9 & 36.3 & 16.3 & 33.4 & 71.3 & 71.5 & 71.3 & 33.8 \\
 \barts & FT & - & - & - & 1e-4 & F & 17.6 & 17.7 & 37.6 & 2.4 & 45.9 & 24.5 & 42.6 & 75.5 & 76.3 & 74.9 & 43.1 \\
 \barts & FT & - & - & - & 1e-4 & F & 17.8 & 18.1 & 38.2 & 2.0 & 46.3 & 25.0 & 43.2 & 75.7 & 76.6 & 75.0 & 43.3 \\
 \barts & FT & L & 1-BS & T & 1e-4 & F & 19.6 & 19.5 & 39.2 & 2.4 & 47.4 & 26.3 & 44.0 & 76.1 & 76.8 & 75.7 & 44.9 \\
 \barts & Noisy & - & - & - & 1e-4 & F & 19.2 & 18.8 & 39.2 & 1.8 & 47.4 & 26.0 & 44.2 & 76.4 & 77.5 & 75.7 & 44.2 \\
 \barts & Att-Rel & - & - & - & 1e-4 & F & 18.8 & 18.6 & 39.1 & 2.0 & 47.3 & 25.9 & 43.9 & 76.1 & 77.0 & 75.4 & 44.2 \\
 \barts & Logits & - & - & - & 1e-4 & F & 19.5 & 19.4 & 39.4 & 1.9 & 47.7 & 26.3 & 44.3 & 76.3 & 77.0 & 75.9 & 45.0 \\
 \barts & Logits & L & 1-BS & T & 1e-4 & F & 20.0 & 19.7 & 39.5 & 1.9 & 47.6 & 26.6 & 44.3 & 76.3 & 76.9 & 75.9 & 45.0 \\
 \barts & Logits & L+U & 1-BS & T & 1e-4 & F & 20.3 & 20.4 & 40.2 & 1.9 & 48.3 & 27.4 & 45.0 & 76.6 & 77.2 & 76.2 & 45.9 \\
 \barts & Logits & L+U & K-BS & T & 1e-4 & F & 20.4 & 20.2 & 39.9 & 1.8 & 47.9 & 27.2 & 44.6 & 76.4 & 77.1 & 76.1 & 45.6 \\
 \barts & Logits & L+U & Samp. & T & 1e-4 & F & 20.4 & 20.4 & 40.5 & 1.8 & 48.7 & 27.6 & 45.2 & 76.7 & 77.3 & 76.3 & 46.2 \\
 \barts & Logits & L+U & H-Samp. & T & 1e-4 & F & 20.4 & 20.2 & 40.1 & 1.7 & 48.3 & 27.3 & 44.9 & 76.7 & 77.4 & 76.3 & 45.8 \\
 \barts & Logits & L+U & Samp. & S & 1e-4 & F & 20.9 & 20.7 & 40.8 & 1.8 & 49.0 & 27.7 & 45.6 & 77.2 & 77.6 & 77.0 & 46.9 \\
 \barts & Logits & L+U & Samp. & T+S & 1e-4 & F & 21.0 & 20.9 & 40.9 & 1.7 & 49.1 & 27.8 & 45.8 & 77.3 & 77.8 & 77.0 & 47.0 \\
 \gptl & FT & - & - & - & 7e-6 & F & 15.0 & 15.8 & 33.3 & 1.7 & 40.3 & 21.7 & 37.8 & 73.2 & 74.7 & 72.0 & 37.9 \\
 \gptm & FT & - & - & - & 5e-4 & F & 11.6 & 12.2 & 27.3 & 3.7 & 33.9 & 16.3 & 31.6 & 69.4 & 70.2 & 68.9 & 32.6 \\
 \gpts & FT & - & - & - & 1e-3 & F & 6.1 & 6.6 & 17.9 & 3.6 & 22.9 & 9.6 & 21.2 & 53.0 & 53.6 & 52.6 & 21.6 \\
 \optm & FT & - & - & - & 1e-4 & F & 10.6 & 10.7 & 25.4 & 2.5 & 30.9 & 16.4 & 28.7 & 52.4 & 53.0 & 52.1 & 29.2 \\
 \opts & FT & - & - & - & 1e-4 & F & 12.6 & 13.4 & 31.2 & 2.1 & 39.2 & 19.2 & 35.3 & 70.1 & 71.2 & 69.6 & 35.1 \\
\bottomrule
\end{tabular}
\end{adjustbox}
\caption{Results for question generation task, \squad{} dataset}
\label{tab:SQuAD}
\end{table*}
\begin{table*}
\centering
\begin{adjustbox}{width=\textwidth}
\begin{tabular}{llllllllllllllllll}
\toprule
 \textbf{Model} & \textbf{Obj.} & \textbf{PT In.} & \textbf{Decoding} & \textbf{PT St.} & \textbf{LR} & \textbf{FT} & \textbf{Dev} & \textbf{BLEU} & \textbf{ROUGE} & \textbf{PPL} & \textbf{R1} & \textbf{R2} & \textbf{RL} & \textbf{BS-F1} & \textbf{BS-P} & \textbf{BS-R} & \textbf{MET} \\
\midrule
  \tfivel & FT & - & - & - & 5e-5 & F & 6.0 & 6.0 & 21.7 & 1.9 & 28.8 & 8.8 & 27.4 & 71.5 & 72.7 & 70.6 & 26.1 \\
 \tfives & FT & - & - & - & 5e-4 & F & 3.7 & 3.6 & 18.1 & 2.5 & 25.2 & 5.4 & 23.7 & 69.4 & 70.3 & 68.7 & 22.4 \\
 \tfives & FT & L & 1-BS & T & 5e-4 & F & 4.0 & 4.2 & 18.5 & 2.8 & 25.4 & 6.1 & 24.0 & 69.3 & 69.8 & 69.0 & 23.1 \\
 \tfives & Noisy & - & - & - & 5e-4 & F & 4.6 & 4.3 & 19.2 & 2.4 & 26.2 & 6.6 & 24.8 & 70.1 & 71.3 & 69.2 & 23.4 \\
 \tfives & Att-Rel & - & - & - & 5e-4 & F & 4.5 & 4.3 & 19.0 & 2.4 & 26.0 & 6.4 & 24.7 & 70.1 & 71.4 & 69.1 & 23.1 \\
 \tfives & Logits & - & - & - & 5e-4 & F & 4.5 & 4.3 & 19.0 & 2.4 & 25.9 & 6.5 & 24.6 & 70.0 & 71.3 & 69.0 & 23.1 \\
 \tfives & Logits & L & 1-BS & T & 5e-4 & F & 4.4 & 4.4 & 19.1 & 2.5 & 26.2 & 6.4 & 24.7 & 69.8 & 70.7 & 69.3 & 23.5 \\
 \tfives & Logits & L+U & 1-BS & T & 5e-4 & F & 4.9 & 4.8 & 19.8 & 2.4 & 26.8 & 7.1 & 25.5 & 70.4 & 71.5 & 69.6 & 24.1 \\
 \tfives & Logits & L+U & K-BS & T & 5e-4 & F & 5.0 & 4.8 & 19.8 & 2.4 & 26.9 & 7.1 & 25.5 & 70.4 & 71.6 & 69.6 & 24.0 \\
 \tfives & Logits & L+U & Samp. & T & 5e-4 & F & 5.0 & 4.7 & 19.8 & 2.4 & 26.8 & 7.0 & 25.5 & 70.6 & 72.0 & 69.4 & 23.8 \\
 \tfives & Logits & L+U & H-Samp. & T & 5e-4 & F & 5.0 & 4.7 & 19.9 & 2.3 & 27.0 & 7.2 & 25.5 & 70.5 & 71.5 & 69.7 & 24.2 \\
 \tfives & Logits & L+U & Samp. & S & 5e-4 & F & 4.9 & 4.7 & 19.7 & 2.3 & 26.7 & 7.0 & 25.4 & 70.5 & 71.9 & 69.5 & 23.8 \\
 \tfives & Logits & L+U & Samp. & T+S & 5e-4 & F & 5.3 & 4.8 & 19.9 & 2.4 & 27.0 & 7.3 & 25.5 & 70.4 & 71.3 & 69.7 & 24.2 \\
 \bartl & FT & - & - & - & 5e-5 & F & 6.4 & 6.0 & 21.4 & 2.1 & 28.5 & 8.6 & 27.1 & 71.5 & 72.7 & 70.6 & 25.7 \\
 \bartb & FT & - & - & - & 5e-5 & F & 4.6 & 4.9 & 20.3 & 2.1 & 27.3 & 7.5 & 26.0 & 71.1 & 72.7 & 69.7 & 24.1 \\
 \bartp & FT & - & - & - & 5e-5 & F & 3.7 & 3.7 & 17.9 & 2.3 & 24.8 & 5.4 & 23.4 & 69.5 & 70.7 & 68.5 & 21.7 \\
 \barts & FT & - & - & - & 5e-5 & F & 3.7 & 3.9 & 18.8 & 2.7 & 25.6 & 6.2 & 24.5 & 70.1 & 72.5 & 68.1 & 22.1 \\
 \barts & FT & L & 1-BS & T & 5e-5 & F & 4.6 & 4.6 & 19.4 & 2.8 & 26.2 & 6.9 & 25.0 & 70.2 & 71.4 & 69.2 & 23.4 \\
 \barts & Noisy & - & - & - & 5e-5 & F & 4.7 & 4.7 & 19.8 & 2.4 & 26.7 & 7.2 & 25.5 & 70.7 & 72.4 & 69.3 & 23.6 \\
 \barts & Att-Rel & - & - & - & 5e-5 & F & 5.1 & 4.4 & 19.2 & 2.7 & 26.1 & 6.8 & 24.8 & 70.4 & 72.0 & 69.1 & 23.0 \\
 \barts & Logits & - & - & - & 5e-5 & F & 5.0 & 4.7 & 19.5 & 2.6 & 26.4 & 7.0 & 25.2 & 70.6 & 72.1 & 69.3 & 23.5 \\
 \barts & Logits & L & 1-BS & T & 5e-5 & F & 5.3 & 5.0 & 19.8 & 2.6 & 26.7 & 7.4 & 25.4 & 70.6 & 71.9 & 69.5 & 23.9 \\
 \barts & Logits & L+U & 1-BS & T & 5e-5 & F & 5.6 & 5.1 & 20.1 & 2.6 & 27.0 & 7.6 & 25.6 & 70.7 & 72.1 & 69.6 & 24.1 \\
 \barts & Logits & L+U & K-BS & T & 5e-5 & F & 5.4 & 5.2 & 20.1 & 2.5 & 27.0 & 7.6 & 25.7 & 70.9 & 72.3 & 69.8 & 24.1 \\
 \barts & Logits & L+U & Samp. & T & 5e-5 & F & 5.6 & 5.2 & 20.2 & 2.5 & 27.2 & 7.7 & 25.8 & 70.9 & 72.2 & 69.8 & 24.3 \\
 \barts & Logits & L+U & H-Samp. & T & 5e-5 & F & 5.3 & 5.0 & 19.9 & 2.5 & 26.9 & 7.4 & 25.6 & 70.8 & 72.3 & 69.6 & 23.8 \\
 \barts & Logits & L+U & Samp. & S & 5e-5 & F & 5.0 & 4.9 & 20.0 & 2.5 & 26.9 & 7.5 & 25.6 & 70.9 & 72.3 & 69.7 & 23.9 \\
 \barts & Logits & L+U & Samp. & T+S & 5e-5 & F & 5.2 & 5.1 & 20.3 & 2.4 & 27.2 & 7.7 & 25.9 & 71.0 & 72.3 & 69.9 & 24.3 \\
 \gptl & FT & - & - & - & 5e-6 & F & 3.6 & 3.6 & 13.8 & 2.3 & 18.5 & 5.1 & 17.6 & 67.2 & 69.4 & 65.4 & 18.9 \\
 \gptm & FT & - & - & - & 5e-4 & F & 1.9 & 2.0 & 9.8 & 4.8 & 13.7 & 2.8 & 12.9 & 63.5 & 64.7 & 62.6 & 15.5 \\
 \gpts & FT & - & - & - & 1e-3 & F & 2.1 & 2.2 & 10.9 & 2.8 & 15.0 & 3.2 & 14.3 & 65.2 & 67.9 & 62.9 & 16.1 \\
 \optm & FT & - & - & - & 1e-4 & F & 2.5 & 3.0 & 15.4 & 3.4 & 21.2 & 4.8 & 20.1 & 61.7 & 62.7 & 60.9 & 19.1 \\
 \opts & FT & - & - & - & 1e-4 & F & 1.9 & 2.1 & 10.9 & 3.6 & 15.5 & 3.0 & 14.3 & 64.1 & 65.0 & 63.7 & 16.0 \\
\bottomrule
\end{tabular}
\end{adjustbox}
\caption{Results for abductive commonsense reasoning, task, \art{} dataset}
\label{tab:ART}
\end{table*}
\begin{table*}
\centering
\begin{adjustbox}{width=\textwidth}
\begin{tabular}{llllllllllllllllll}
\toprule
 \textbf{Model} & \textbf{Obj.} & \textbf{PT In.} & \textbf{Decoding} & \textbf{PT St.} & \textbf{LR} & \textbf{FT} & \textbf{Dev} & \textbf{BLEU} & \textbf{ROUGE} & \textbf{PPL} & \textbf{R1} & \textbf{R2} & \textbf{RL} & \textbf{BS-F1} & \textbf{BS-P} & \textbf{BS-R} & \textbf{MET} \\
\midrule
\tfivel & FT & - & - & - & 5e-5 & F & 25.4 & 25.7 & 45.4 & 1.5 & 54.0 & 31.5 & 50.6 & 78.4 & 78.4 & 78.6 & 53.8 \\
 \tfives & FT & - & - & - & 5e-4 & F & 23.3 & 23.3 & 43.4 & 2.1 & 52.3 & 29.0 & 48.9 & 76.9 & 76.9 & 77.0 & 51.6 \\
 \tfives & FT & L & 1-BS & T & 5e-4 & F & 24.2 & 23.7 & 44.1 & 2.7 & 53.1 & 29.7 & 49.5 & 77.4 & 77.2 & 77.7 & 52.5 \\
 \tfives & Noisy & - & - & - & 5e-4 & F & 24.5 & 24.2 & 44.3 & 1.9 & 53.1 & 30.1 & 49.7 & 77.5 & 77.6 & 77.5 & 52.5 \\
 \tfives & Att-Rel & - & - & - & 5e-4 & F & 24.5 & 24.1 & 44.4 & 1.9 & 53.3 & 30.0 & 49.8 & 77.4 & 77.4 & 77.5 & 52.5 \\
 \tfives & Logits & - & - & - & 5e-4 & F & 24.1 & 24.3 & 44.5 & 1.9 & 53.4 & 30.2 & 50.0 & 77.4 & 77.4 & 77.5 & 52.7 \\
 \tfives & Logits & L & 1-BS & T & 5e-4 & F & 24.8 & 24.8 & 44.7 & 1.9 & 53.4 & 30.5 & 50.2 & 77.6 & 77.7 & 77.6 & 52.8 \\
 \tfives & Logits & L+U & 1-BS & T & 5e-4 & F & 25.5 & 25.1 & 45.4 & 1.9 & 54.2 & 31.3 & 50.7 & 78.0 & 78.0 & 78.1 & 53.7 \\
 \tfives & Logits & L+U & K-BS & T & 5e-4 & F & 25.5 & 25.4 & 45.5 & 1.8 & 54.2 & 31.3 & 51.0 & 78.1 & 78.1 & 78.2 & 53.7 \\
 \tfives & Logits & L+U & Samp. & T & 5e-4 & F & 25.4 & 25.4 & 45.5 & 1.8 & 54.2 & 31.2 & 50.9 & 78.1 & 78.2 & 78.1 & 53.5 \\
 \tfives & Logits & L+U & H-Samp. & T & 5e-4 & F & 25.5 & 25.5 & 45.1 & 1.7 & 53.9 & 31.0 & 50.6 & 78.1 & 78.1 & 78.1 & 53.4 \\
 \tfives & Logits & L+U & Samp. & S & 5e-4 & F & 25.2 & 25.5 & 45.4 & 1.8 & 54.1 & 31.1 & 50.9 & 78.1 & 78.1 & 78.2 & 53.7 \\
 \tfives & Logits & L+U & Samp. & T+S & 5e-4 & F & 25.3 & 25.7 & 45.3 & 1.7 & 54.0 & 31.2 & 50.8 & 78.1 & 78.2 & 78.2 & 53.6 \\
 \bartl & FT & - & - & - & 5e-5 & F & 25.3 & 25.1 & 44.8 & 1.8 & 53.4 & 30.7 & 50.2 & 78.3 & 78.5 & 78.2 & 52.9 \\
 \bartb & FT & - & - & - & 5e-5 & F & 25.1 & 23.8 & 43.7 & 1.8 & 52.3 & 29.5 & 49.1 & 77.4 & 77.6 & 77.3 & 51.6 \\
 \bartp & FT & - & - & - & 5e-5 & F & 22.7 & 22.0 & 41.8 & 2.0 & 50.7 & 27.4 & 47.3 & 76.1 & 76.4 & 75.9 & 49.5 \\
 \barts & FT & - & - & - & 5e-5 & F & 23.1 & 22.4 & 43.0 & 2.6 & 51.7 & 28.6 & 48.6 & 76.7 & 77.3 & 76.3 & 50.0 \\
 \barts & FT & L & 1-BS & T & 5e-5 & F & 24.8 & 23.3 & 43.6 & 3.0 & 52.6 & 29.1 & 49.2 & 77.2 & 77.4 & 77.2 & 51.5 \\
 \barts & Noisy & - & - & - & 5e-5 & F & 24.1 & 23.1 & 44.0 & 2.1 & 52.7 & 29.7 & 49.5 & 77.3 & 77.9 & 76.8 & 51.2 \\
 \barts & Att-Rel & - & - & - & 5e-5 & F & 23.3 & 22.2 & 43.2 & 2.3 & 52.0 & 28.8 & 48.7 & 76.8 & 77.4 & 76.3 & 50.3 \\
 \barts & Logits & - & - & - & 5e-5 & F & 24.2 & 23.8 & 44.1 & 2.4 & 52.9 & 29.8 & 49.5 & 77.5 & 77.8 & 77.2 & 51.8 \\
 \barts & Logits & L & 1-BS & T & 5e-5 & F & 25.0 & 23.8 & 44.1 & 2.4 & 53.0 & 29.7 & 49.6 & 77.5 & 77.8 & 77.4 & 51.9 \\
 \barts & Logits & L+U & 1-BS & T & 5e-5 & F & 25.2 & 24.4 & 44.8 & 2.3 & 53.6 & 30.5 & 50.3 & 77.8 & 78.1 & 77.6 & 52.7 \\
 \barts & Logits & L+U & K-BS & T & 5e-5 & F & 25.6 & 24.5 & 44.7 & 2.2 & 53.3 & 30.5 & 50.1 & 77.7 & 78.0 & 77.5 & 52.1 \\
 \barts & Logits & L+U & Samp. & T & 5e-5 & F & 25.5 & 24.7 & 45.0 & 2.1 & 53.7 & 30.8 & 50.4 & 77.9 & 78.3 & 77.7 & 52.6 \\
 \barts & Logits & L+U & H-Samp. & T & 5e-5 & F & 25.2 & 24.7 & 45.0 & 2.0 & 53.7 & 30.8 & 50.5 & 77.8 & 78.2 & 77.6 & 52.6 \\
 \barts & Logits & L+U & Samp. & S & 5e-5 & F & 25.5 & 24.5 & 44.8 & 2.1 & 53.6 & 30.5 & 50.3 & 77.9 & 78.2 & 77.7 & 52.5 \\
 \barts & Logits & L+U & Samp. & T+S & 5e-5 & F & 26.1 & 24.8 & 45.2 & 2.0 & 53.9 & 30.9 & 50.7 & 78.1 & 78.3 & 78.0 & 53.2 \\
 \gptl & FT & - & - & - & 1e-5 & F & 21.6 & 20.3 & 39.0 & 1.8 & 47.4 & 25.5 & 44.2 & 74.5 & 75.7 & 73.5 & 45.7 \\
 \gptm & FT & - & - & - & 7e-4 & F & 18.6 & 17.7 & 35.0 & 4.3 & 42.9 & 22.4 & 39.6 & 70.6 & 70.7 & 70.7 & 42.3 \\
 \gpts & FT & - & - & - & 1e-3 & F & 17.6 & 17.6 & 34.1 & 2.2 & 41.6 & 22.2 & 38.6 & 65.3 & 65.9 & 65.0 & 41.0 \\
 \optm & FT & - & - & - & 3e-4 & F & 19.0 & 18.9 & 38.1 & 3.6 & 46.6 & 24.4 & 43.2 & 72.1 & 72.8 & 71.5 & 44.7 \\
 \opts & FT & - & - & - & 5e-5 & F & 20.8 & 20.4 & 40.5 & 1.9 & 49.1 & 26.7 & 45.7 & 74.6 & 75.4 & 74.0 & 46.3 \\
\bottomrule
\end{tabular}
\end{adjustbox}
\caption{Results for style transfer and simplification task, \shakespeare{} dataset}
\label{tab:Shakespeare}
\end{table*}

\clearpage
\section{Human Evaluation for \art{}}
\label{sec:human_art}

\begin{table}
\large
\centering
\begin{adjustbox}{width=0.4\textwidth}
\begin{tabular}{ccccc}
\toprule
 \textbf{References} & \textbf{\tfives} & \textbf{\tfivel} & \textbf{\tfivekd} & \textbf{Gap} \\
\midrule
4.45 & 2.75 & 4.0 & 3.65 & 72\% \\
\bottomrule
\end{tabular}
\end{adjustbox}
\caption{Human evaluation for the \art{} dataset. The numbers are the average ratings for the golden references, the student baseline (\tfives), the teacher (\tfivel), and the final distilled model (\tfivekd). We also present the fraction of the student-teacher gap closed by \tfivekd.}
\label{tab:human_art}
\vspace{-1em}
\end{table}

In this section, we aim is to examine the relatively lower performance of our KD method on the \art{} dataset when compared to other datasets. Accordingly, the abductive reasoning task and the \art{} dataset in particular, are a unique case where automatic evaluation is hard to perform due to the large number of diverse potential solutions (See the examples in \S\ref{sub:examples}). Therefore, we assume that the fraction of the student-teacher gap closed by the distilled model may have been underestimated. 

To validate our assumption, we conducted a human evaluation involving two annotators. We randomly selected 50 input examples and generated outputs using the student baseline (\tfives), the teacher (\tfivel), and the final distilled model (\tfivekd). The annotators were asked to rate the generated texts on a five-level scale (see \S\ref{sub:instructions} below). The inter-annotator agreement achieved by our annotators was Kendall's $\tau$=0.52. According to Table~\ref{tab:human_art}, which presents the average rating for each model, we find that the distilled model closes 72\% of the student-teacher gap. This result is much greater than the 50\% estimated by the automatic evaluation, and is in-line with the performance of the distilled model on other datasets.

\subsection{Human Evaluation -- Instructions}
\label{sub:instructions}

You will be presented with two texts referred to as ``The Observations''. These observations occur in a specific order, with the first happening before the second. Your task is to assess four ``explanations'' of the second observation, which aim to explain why it occurred, given that the first observation had already occurred.
A good ``explanation'' should be clear and provide a plausible account of what happened between the two observations. You should rate each explanation using this five-level scale:
\begin{enumerate}
    \item The explanation is nonsensical or contains many grammatical errors.
    \item The explanation is not related to the observations or repeating the observations.
    \item The explanation is related to the observations but does not explain them.
    \item The explanation is related to the observations but only partially explains them.
    \item The explanation fully explains the observations.
\end{enumerate}

\subsection{Generated Examples}
\label{sub:examples}

\noindent\textbf{Observation 1:} I went to a rap show for the first time in my life.\\
\noindent\textbf{Observation 2:} Now I'm avid rap and hip hop listener.\\
\noindent\textbf{Reference:} I really enjoyed the show.\\
\noindent\textbf{\tfives :} I went to a hip hop show.\\
\noindent\textbf{\tfivel :} I fell in love with rap and hip hop.\\
\noindent\textbf{\tfivekd :} The rap show was very good.\\

\noindent\textbf{Observation 1:} Allison wanted to renew her vows with Tom.\\
\noindent\textbf{Observation 2:} Yeah even had a new baby.\\
\noindent\textbf{Reference:} Allison and Tom did it and felt more love.\\
\noindent\textbf{\tfives :} Tom had a baby.\\
\noindent\textbf{\tfivel :} Allison proposed to Tom.\\
\noindent\textbf{\tfivekd :} Allison asked Tom to marry her.\\

\noindent\textbf{Observation 1:} Today I decided to learn how to make bread.\\
\noindent\textbf{Observation 2:} I noticed I made an error in my measurements and started over.\\
\noindent\textbf{Reference:} I accidentally put in twice as much salt as needed.\\
\noindent\textbf{\tfives :} I went to the grocery store to learn how to make bread.\\
\noindent\textbf{\tfivel :} I didn't follow the recipe exactly.\\
\noindent\textbf{\tfivekd :} I did not follow the instructions carefully.\\

\noindent\textbf{Observation 1:} Tommy called on the girl who sat next to him in class.\\
\noindent\textbf{Observation 2:} Tommy decided to ask the girl for a date.\\
\noindent\textbf{Reference:} The girl was very beautiful and kind.\\
\noindent\textbf{\tfives :} Tommy asked the girl for a date.\\
\noindent\textbf{\tfivel :} Tommy liked the girl a lot.\\
\noindent\textbf{\tfivekd :} The girl said she liked Tommy.\\

\clearpage
\onecolumn

\end{document}